\documentclass{applemlr}
\usepackage{hyperref}
\usepackage{url}
\usepackage{graphicx}
\usepackage{multirow}
\usepackage{booktabs}
\usepackage{siunitx}
\usepackage{graphicx}
\usepackage{algorithm}
\usepackage{algpseudocode}
\usepackage{amsmath}
\usepackage{cleveref}
\usepackage{enumitem}
\usepackage{xcolor}
\usepackage{wrapfig}
\usepackage{listings}
\usepackage{subcaption}
\usepackage[T1]{fontenc}
\usepackage{xcolor}
\usepackage{amsthm}
\usepackage{listings}
\usepackage{adjustbox}
\usepackage[svgnames]{xcolor} 
\usepackage[normalem]{ulem}   
\usepackage{amsmath}
\usepackage{enumerate}
\usepackage{algorithm}
\usepackage{algpseudocode}
\usepackage{amsfonts}
\usepackage{amsthm}
\usepackage{newtxtt}
\usepackage{cleveref}
\usepackage{diagbox}
\usepackage{colortbl}
\usepackage{amssymb}
\usepackage{xspace}
\usepackage{wrapfig}
\usepackage{adjustbox}
\usepackage{tabularx}
\usepackage{booktabs}
\usepackage{mathtools}
\usepackage{tikz}
\usepackage{enumitem}
\usepackage{silence}
\usepackage{dsfont}
\usepackage[table]{xcolor}
\usepackage[dvipsnames]{xcolor}
\usepackage{multirow}
\usepackage{makecell}
\usepackage{xfakebold}
\usepackage{wrapfig}
\usepackage{mdframed} 
\usepackage{newfloat}

\usepackage{amsthm}
\usepackage{listings}
\usepackage[textwidth=18mm]{todonotes}
\usepackage{float}

\usepackage{amsmath,amsfonts,bm}

\def\eqref#1{equation~\ref{#1}}

\def\1{\bm{1}}

\DeclareMathAlphabet{\mathsfit}{\encodingdefault}{\sfdefault}{m}{sl}
\SetMathAlphabet{\mathsfit}{bold}{\encodingdefault}{\sfdefault}{bx}{n}

\newmdenv[
  backgroundcolor=gray!2,                  
  linecolor=gray!20,                       
  linewidth=0.5pt,                         
  roundcorner=5pt,                         
  font=\sffamily,                          
  frametitlefont=\sffamily\bfseries,       
  frametitlerule=false,                    
  frametitlealignment=\center,             
  innertopmargin=1em,                      
  innerbottommargin=1em,                   
  skipabove=1em,                           
  skipbelow=1em,                           
]{mymessagebox}

\DeclareFloatingEnvironment[
  fileext=lop,
  listname={List of Prompts},
  name=Prompt,
  placement=htp,
  within=none, 
]{prompt}

\lstdefinestyle{python-diff}{
  language=Python,
  basicstyle=\ttfamily\scriptsize,
  keywordstyle=\color{blue},
  stringstyle=\color{purple},
  commentstyle=\color{gray},
  showstringspaces=false,
  escapechar=|,
  breaklines=true,
  breakatwhitespace=true,
  frame=single,
  numbers=left,
  numberstyle=\tiny\color{gray},
   literate={%
    {[}{{[}}1
    {]}{{]}}1
    {\{}{{{\{}{}}}1
    {\}}{{{\}}}}1}
}

\newcommand{\add}[1]{\colorbox{green!20}{\texttt{#1}}}
\newcommand{\del}[1]{\colorbox{red!20}{\texttt{\sout{#1}}}}

\definecolor{textgray}{HTML}{6E6E73}
\usetikzlibrary{positioning, calc}
\usetikzlibrary{decorations.pathmorphing}

\makeatletter
\patchcmd{\wrong@fontshape}{\@gobbletwo}{}{}{}
\makeatother
\numberwithin{equation}{section}
\tcbuselibrary{minted}
\setminted[python]{
  linenos,
  breaklines,
  fontsize=\footnotesize,
  xleftmargin=2em
}

\makeatletter
\AtBeginDocument{
  \urlstyle{sf}
  
}
\makeatother

\definecolor{light}{RGB}{125, 125, 125}
\crefname{tcb@cnt@pbox}{code}{code}
\Crefname{tcb@cnt@pbox}{Code}{Code}
\crefname{assumption}{assumption}{assumption}
\Crefname{assumption}{Assumption}{Assumptions}

\newtcolorbox[auto counter]{pbox}[2][]{
  colback=white,
  title=Code~\thetcbcounter: #2,
  #1,fonttitle=\sffamily,
  fontupper=\sffamily,
  arc=2pt,
  colframe=bgcolor,
  coltitle=fgcolor,
  colbacktitle=bgcolor,
  toptitle=0.25cm,
  bottomtitle=0.125cm
}

\makeatletter
\newcommand\applefootnote[1]{%
  \begingroup
  \renewcommand\thefootnote{}%
  \renewcommand\@makefntext[1]{\noindent##1}%
  \footnote{#1}%
  \addtocounter{footnote}{-1}%
  \endgroup
}
\makeatother

\definecolor{cverbbg}{gray}{0.90}

\title{GRACE: A Language Model Framework for Explainable Inverse Reinforcement Learning}

\abstract{
Inverse Reinforcement Learning aims to recover reward models from expert demonstrations, but traditional methods yield black-box models that are difficult to interpret and debug. In this work, we introduce \method{} (\textbf{G}enerating \textbf{R}ewards \textbf{A}s \textbf{C}od\textbf{E}), a method for using Large Language Models within an evolutionary search to reverse-engineer an interpretable, code-based reward function directly from expert trajectories. The resulting reward function is executable code that can be inspected and verified. 
We empirically validate \method{} on the MuJoCo, BabyAI and AndroidWorld benchmarks, where it efficiently learns highly accurate rewards, even in complex, multi-task settings. Further, we demonstrate that the resulting reward leads to strong policies, compared to both competitive Imitation Learning and online RL approaches with ground-truth rewards.
Finally, we show that \method{} is able to build complex reward APIs in multi-task setups.}

\metadata[Correspondence]{\sffamily Silvia Sapora: \url{silvia.sapora@stats.ox.ac.uk}; Bogdan Mazoure: \url{bmazoure@apple.com}}
\date{\sffamily\today}

\author{Silvia Sapora$^*$}
\author{Devon Hjelm}
\author{Omar Attia}
\author{Alexander Toshev}
\author{Bogdan Mazoure}

\affiliation{Apple}

\newcommand{\method}{GRACE}

\begin{document}

\maketitle

\applefootnote{$^*$Work done while SS was an intern at Apple.}

\section{Introduction}
The performance of modern Reinforcement Learning (RL) agents is determined by, among other factors, the quality of their reward function. Traditionally, reward functions are defined manually as part of the problem specification. In many real-world settings, however, environments are readily available while reward functions are absent and must be specified. Manually designing rewards is often impractical, error-prone, and does not scale, particularly in contemporary multi-task RL scenarios~\citep{wilson2007multi, teh2017distral, parisotto2015actor}.

A natural alternative is to automate reward specification by learning a reward model from data. The dominant paradigm here is Inverse Reinforcement Learning (IRL), which attempts to infer a reward model from observations of expert behavior~\citep{ng2000irl, christiano2017preferences, ziebart2008maximum}. In the era of Deep RL, approaches such as AIRL~\citep{fu2018learningrobustrewardsadversarial} represent rewards with deep neural networks. While effective, these reward functions are typically opaque black boxes, making them difficult to interpret or verify~\citep{molnar2020interpretable}. Moreover, IRL methods often require substantial amounts of data and can lead to inaccurate rewards~\citep{sapora2024evil}.

Recently, expressing reward models through code has emerged as a promising approach ~\citep{pmlr-v235-venuto24a,ma2023eureka}. Code is a particularly well-suited representation, as reward functions are often far simpler to express than the complex policies that maximize them ~\citep{ng2000irl,cookcomplexity, godel}. These approaches leverage code-generating Large Language Models (LLMs) and human-provided task descriptions or goal states to generate reward programs~\citep{pmlr-v235-venuto24a}. Subsequently, the generated rewards are verified~\citep{pmlr-v235-venuto24a} or improved using the performance of a trained policy as feedback~\citep{ma2023eureka}. However, this prior work has not investigated whether it is possible to recover a reward function purely from human demonstrations in an IRL-style setting, without utilizing any explicit task description or domain-specific design assumptions.

In this work, we address the question of how to efficiently infer rewards-as-code from expert demonstrations using LLMs. We propose an optimization procedure inspired by evolutionary search~\citep{goldberg1989geneticalgorithms, eiben2003evolutionary, salimans2017es, romera2024funsearch, novikov2025alphaevolvecodingagentscientific}, in which code LLMs iteratively introspect over demonstrations to generate and refine programs that serve as reward models. This perspective effectively revisits the IRL paradigm in the modern context of program synthesis with LLMs.

Our contributions are threefold:
\begin{itemize}
    \item \textbf{Sample-Efficient Reward Recovery:} We demonstrate that code LLMs conditioned on expert demonstrations can produce highly accurate reward models that generalize well to held-out data. Crucially, \method{} is highly sample-efficient, recovering accurate rewards from relatively few demonstrations without requiring manual domain knowledge or human-in-the-loop guidance.
    \item \textbf{Well-Shaped Rewards:} We show that rewards learned by \method{} enable the training of strong policies across diverse domains, including the procedural \emph{BabyAI}, continuous control in \emph{MuJoCo}, and real-world device control in \emph{AndroidWorld}. Empirical results indicate that \method{} matches or outperforms established IRL baselines (e.g., GAIL) and online RL with ground-truth, sparse rewards.
    \item \textbf{Interpretability and Modularity:} Unlike black-box neural networks, \method{} generates rewards as executable Python code, making them inherently interpretable and verifiable. Furthermore, in multi-task settings, our evolutionary search naturally emerges a modular library of reusable reward functions, enabling efficient generalization across tasks.
\end{itemize}

\section{Related Works}
\textbf{LLMs for Rewards} A common way to provide verification/reward signals in an automated fashion is to utilize Foundation Models. LLM-based feedback has been used directly by \cite{zheng2023judging} to score a solution or to critique examples~\citep{zankner2024critique}. Comparing multiple outputs in a relative manner has been also explored by~\cite{wang2023llm}. Note that such approaches use LLM in a zero shot fashion with additional prompting and potential additional examples. Hence, they can utilize only a small number of demonstrations at best. In addition to zero shot LLM application, it is also common to train reward models, either from human feedback~\citep{ouyang2022training} or from AI feedback~\citep{klissarov2023motif,klissarov2024modeling}. However, such approaches require training a reward model that isn't interpretable and often times require a larger number of examples.

\textbf{Code as Reward} As LLMs have emerged with powerful program synthesis capabilities ~\citep{chen2021codex,austin2021programsynthesis,li2023starcoder,fried2022incoder,nijkamp2022codegen} research has turned towards generating environments for training agents ~\citep{zala2024envgen,faldor2025omni-epic} for various domains and complexities. 

When it comes to rewards in particular, code-based verifiers use a language model to generate executable Python code based on a potentially private interface such as the environment's full state. Because early language models struggled to reliably generate syntactically correct code, the first code-based verifiers ~\citep{yu2023language,venuto2024code} implemented iterative re-prompting and fault-tolerance strategies. More recent approaches focus on progressively improving a syntactically correct yet suboptimal reward function, particularly by encouraging exploration ~\citep{romera2024mathematical,novikov2025alphaevolve}. 
Other approaches such as~\cite{zhou2023language, dainese2024generating} use search in conjunction with self-reflection~\citep{madaan2023self} to provide feedback.

While closely related to our work, EUREKA \citep{ma2023eureka} assumes access to the ground-truth reward signal to evaluate and evolve the generated code, while Reward-As-Code \citep{venuto2024code} depends on explicit task descriptions or goal states and a hand written pipeline. In contrast to these earlier works, our pipeline doesn't require any domain-specific adaptation and does not require the ground truth reward signal or a description of the task.

\textbf{Inverse Reinforcement Learning (IRL)} Early approaches infer a reward function that makes the expert's policy optimal over all alternatives~\citep{ng2000irl}. Subsequent works focused on learning policies directly, without explicit reward recovery~\citep{abbeel2004apprenticeship}, while incorporating entropy regularization~\citep{ziebart2008maximum} or leveraging convex formulations~\citep{ratliff2006maximum}. More recently, deep IRL methods such as Generative Adversarial Imitation Learning (GAIL)~\citep{ho2016gail} have framed the problem as a distribution matching task using adversarial training. Adversarial Inverse Reinforcement Learning (AIRL)~\citep{fu2018learningrobustrewardsadversarial} further improves upon this by recovering disentangled and transferable reward functions. While related to our formulation, our representation (code) and our optimization strategy (evolutionary search) are fundamentally different, as \method{} benefits from implicit regularization through its symbolic reward representation.

\section{Background}
\paragraph{Reinforcement Learning}
We consider a finite-horizon Markov Decision Process (MDP)~\citep{puterman2014markov} parameterized by $\mathcal{M}=\langle \mathcal{S}, \mathcal{A}, T, r \rangle$ where $\mathcal{S}$, $\mathcal{A}$ are the state and action spaces, $T: \mathcal{S} \times \mathcal{A}\rightarrow \Delta(\mathcal{S})$ is the transition operator, and $r$ is a reward function. The agent's behavior is described by the policy $\pi: \mathcal{S} \to \Delta(\mathcal{A})$. Starting from a set of initial states $\mathcal{S}_0\subset \mathcal{S}$, the agent takes the action $a\sim \pi(s)$ at $s$, receives a reward $r(s)$ and transitions into state $s'\sim T(s,a)$. 

The performance of the agent is measured with expected cumulative per-timestep rewards, referred to as return:
\begin{equation}
    J(\pi, r)=\mathbb{E}_{\tau \sim \pi, T}[\sum_{t=1}^H r(s_t)]
    \label{eq:rl_objective}
\end{equation}
where $\tau$ are trajectory unrolls of horizon $H$ of the policy $\pi$ in $\mathcal{M}$. An optimal agent can be learned by maximizing~\Cref{eq:rl_objective} via gradient descent with respect to the policy, also known as policy gradient~\citep{sutton1999policy,schulman2017proximal}.
\paragraph{Inverse Reinforcement Learning} If the reward $r$ is not specified, it can be learned from demonstrations of an expert policy $\pi_E$. In particular, the classical IRL objective learns a reward whose optimal return is attained by the expert~\citep{ng2000irl,syed2007game}:
\begin{equation}
\label{eq:irl}
\max_R \min_\pi J(\pi_E, r) - J(\pi, r)
\end{equation}
More recent IRL approaches learn a discriminator that distinguishes between expert and non-expert demonstrations~\citep{ho2016gail, swamy2021moments}. The likelihood of the agent's data under the trained discriminator can be implicitly thought of as a reward. Modern approaches often frame this as a divergence minimization problem, matching the state-action visitation distributions of the learner and the expert. These approaches utilize gradient based methods to optimize their objectives.




\paragraph{Evolutionary search} As an alternative for cases where the objective is not readily differentiable, gradient-free methods can be employed. One such method is evolutionary search, which maintains a set of candidate solutions (called a population) and applies variation operators to improve it ~\citep{salimans2017evolutionstrategiesscalablealternative, evolutionarycomputing, geneticalgorithmsearch}. These operators include mutation, where a hypothesis is partially modified, and recombination, where two hypotheses are combined to produce a new one. Each variation is evaluated using a fitness function, which measures the quality of a given hypothesis. Starting with an initial population, evolutionary search repeatedly applies these variation operators, replacing hypotheses with higher-fitness alternatives.

In this work, we focus on inferring reward functions, represented as Python code, from a set of demonstrations. While this setup is related to 
IRL, representing rewards as code prevents us from applying gradient-based methods commonly used in IRL. For this reason, we adopt evolutionary search as our optimization method.

\section{Method}
\begin{figure*}[ht!]
    \centering
    \includegraphics[width=0.95\columnwidth,trim={0cm 27.5cm 9cm 35cm}, clip]{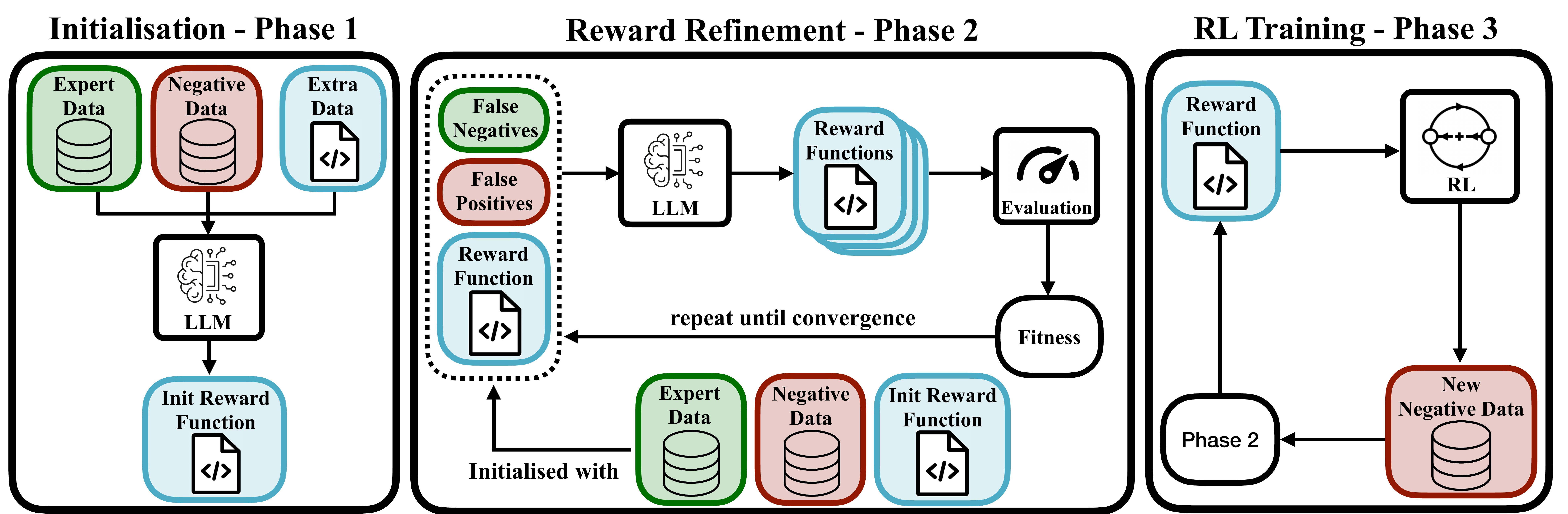} 
    \caption{Overview of the \method{} framework. \textbf{(a)} The expert, negative and extra data (if any) is used to generate an initial set of possible reward functions. \textbf{(b)} The expert and negative states are used to mutate reward functions through an evolutionary procedure. The rewards are iteratively refined by feeding low-fitness examples to the reward. \textbf{(c)} An agent is trained with online RL using the converged reward; the data it sees during the training is added to $\mathcal{D}^{-}$ and used to further improve the reward.}
    \label{fig:method_overview}
\end{figure*}

\begin{algorithm*}[t]
\caption{\method{}: Generating Rewards As CodE}
\begin{minipage}[t]{0.48\linewidth}
\begin{algorithmic}
\State \textbf{Inputs:} 
\State $\mathcal{D}^+$: expert trajectories, $\mathcal{D}^-$: random trajectories
\State $\mathcal{R}$: reward population
\State \textbf{Parameters:} 
\State $P$: population size, $N$: RL budget
\State $M$: data augmentation steps
\Statex
\Procedure{\method{}($\mathcal{D}^+$, $\mathcal{D}^-$)}{}
    \State $\mathcal{S}_e=\{s\in D^+\}$, $\mathcal{S}_n=\{s\in D^-\}$
        \State $\mathcal{R} = \{\textrm{LLM}(S_e, S_{n}, \textrm{reward\_prompt})\}$
    \item[]
    \State // Reward Refinement.
    \For{$i=1\dots M$}
        \State $\mathcal{R} \gets$ \textsc{EvoSearch}($\mathcal{R},\mathcal{S}_e,\mathcal{S}_{n}$)
        \State $\mathcal{D}_{r^*} \gets$ 
        \textsc{DataExpandRL}($\mathcal{R}$)
        \State $\mathcal{S}_{n} = \mathcal{D}_{r^*} \cup \mathcal{S}_{n}$
    \EndFor
    \State \textbf{Return} $r^*=\arg\max_{r\in\mathcal{R}} f(r)$
\EndProcedure
\end{algorithmic}
\end{minipage}\hfill
\begin{minipage}[t]{0.48\linewidth}
\begin{algorithmic}
\Function{\textsc{EvoSearch}}{($\mathcal{R}, \mathcal{S}_e, \mathcal{S}_{n}$)}
    \For{$k=1\dots K$}
        \State $\mathcal{R}_{\text{new}} \gets \emptyset$
        \For{$j=1\dots P$}
            \State Sample $r \sim \exp(f(r))$
            \State $r' \gets m(r)$ // See Eq.~\ref{eq:mutation}
            \State $\mathcal{R}_{\text{new}} \gets \mathcal{R}_{\text{new}} \cup \{r'\}$
        \EndFor
        \State $\mathcal{R} \gets \text{Top}_P(\mathcal{R} \cup \mathcal{R}_{\text{new}})$ // Keep best $P$ rewards
    \EndFor
    \State \textbf{Return} $\mathcal{R}$
\EndFunction
\Statex
\Function{DataExpandRL}{$\mathcal{R}$}
    \State $r^* \gets \arg\max_{r\in\mathcal{R}} f(r)$
    \State Train $\pi_{r^*}$ with PPO under budget $N/M$
    \State Collect new trajectories $\mathcal{D}_{r^*}$
    \State \textbf{return} $\mathcal{D}_{r^*}$
\EndFunction
\end{algorithmic}
\end{minipage}
\label{alg:grace}
\end{algorithm*}

We propose \method{}, \textbf{G}enerating \textbf{R}ewards \textbf{A}s \textbf{C}od\textbf{E}, an interpretable IRL framework that generates a reward function as executable Python code. 
Initially, an LLM analyzes expert and random trajectories (Phase 1) and generates a preliminary set of reward programs. This initial set of reward functions is then iteratively improved through evolutionary search, where the LLM mutates the code based on low-fitness examples (Phase 2). The best-performing reward function is used to train an RL agent. This agent then explores the environment, and the new trajectories it generates are used to expand the dataset, revealing new edge cases or failure modes (Phase 3). This loop continues, with the expanded dataset from Phase 3 being used in the next iteration of Phase 1 and 2, progressively improving the reward function. Notably, prior to the initial analysis, the LLM can optionally perform a data cleaning step to identify and remove irrelevant states if the expert trajectories are known to be noisy or suboptimal.
The overall process is illustrated in~\Cref{fig:method_overview} and detailed below and in Algorithm~\ref{alg:grace}



\paragraph{Phase 1: Initialisation}\label{sec:init}
The initial reward code generation by \method{} is based on a set of demonstration trajectories $\mathcal{D^{+}}$ and a set of random trajectories $\mathcal{D}^{-}$. The former is generated using an expert policy or human demonstrations depending on the concrete setup, while the latter is produced by a random policy. Note that with a slight abuse of notation we will use $\mathcal{D}$ to denote interchangeably a set of trajectories as well the set of all states from these trajectories.

The language model is prompted with a random subset of $\mathcal{D}^{+}$ and, optionally, extra information available about the environment (e.g. its Python code or tool signature), to produce two artifacts:

\paragraph{Initial rewards:} The LLM generates an initial set $\mathcal{R}^\mathrm{init}$ of reward functions. Each function $r\in\mathcal{R}^\mathrm{init}$ is represented as Python code:
\begin{verbatim}
def reward(state) -> float:
    <LLM produced code>
\end{verbatim}
designed to assign high values to expert states $\mathcal{S}_e$ and low values to negative ones $\mathcal{S}_{n}$. This set of rewards is treated as the population in the subsequent evolution phase.
\paragraph{(Optional) Data cleaning:} The LLM analyzes the states from expert demonstrations to identify the subset of expert states $\mathcal{S}_e \subseteq \mathcal{D}^{+}$ that solve the task - these are positive samples. All remaining states $\mathcal{S}_{n}=\{\mathcal{D}^{+} \setminus \mathcal{S}_e\} \cup \mathcal{D}^{-}$ are treated as negative samples.


\paragraph{Phase 2: Reward Refinement through Evolutionary Search} \method{} uses Evolutionary Search to discover rewards that effectively distinguish between expert and negative states. This is achieved by \emph{mutating} the current reward population $\mathcal{R}$ using a code LLM and retaining rewards with high \emph{fitness}.

The \emph{fitness} $f$ of a reward function $r$ measures how well this function assigns large values to expert states and small values to negative states, akin to what would be expected from a meaningful reward. To do this, we adopt the standard AIRL \citep{fu2018learningrobustrewardsadversarial} loss to ensure transferable rewards:
\begin{equation}
f(r) = \mathbb{E}_{s \sim \mathcal{S}e}[\log D_r(s)] + \mathbb{E}_{s \sim \mathcal{S}_n}[\log(1 - D_r(s))] 
\label{eq:fitness_grace}
\end{equation}
where $D_r(s)$ is the discriminator parametrised by the reward function $r$. Following the AIRL formulation, the discriminator takes the form $D_r(s) = \frac{exp(r(s))}{exp(r(s))+\pi(a|s)}$.

The \emph{mutation} operator $m$ utilizes an LLM to improve the current reward code and correct failures. We prompt the LLM to generate a mutated reward $r'$ based on the current source code and specific failure cases:
\begin{equation}\label{eq:mutation}
m(r) = \textrm{LLM}(\textrm{source}(r), \textrm{context}, \textrm{prompt})
\end{equation}
Crucially, the $\textrm{context}$ provided to the LLM enables targeted debugging. It includes:
\begin{itemize}
    \item \textbf{Current Reward Source Code:} The Python implementation of the ``parent'' reward function $r$.
    \item \textbf{Wrong Examples:} A random sample of states $s_w$ with low fitness under reward $r$.
    \item \textbf{Reward Values:} The output value $r(s_w)$ by the current function for each low-fitness state.
    \item \textbf{(Optional) Extra environment information:} It is possible to include, if available, extra information about the environment such as (partial) source code or a text description.
    \item \textbf{(Optional) Debug information:} The LLM can have an option to define a $\texttt{debug}(s, D^+, D^-)$ function. The output of this function can then be included, in-context, in each mutation prompt. This allows the model to print custom intermediate variables, logic checks during execution or aggregate information over the expert and negative datasets.
\end{itemize}


We repeatedly apply the mutation operation to evolve the reward population $\mathcal{R}$. In each iteration, we sample a set of parent rewards $r \in \mathcal{R}$ based on a softmax distribution of their fitness, where the probability of selecting a specific reward is given by $\frac{\exp(f(r))}{\sum_{r' \in \mathcal{R}} \exp(f(r'))}$. We apply the mutation operator to these parents to generate a new batch of candidate rewards. To maintain a high-quality population, we retain only the top $N$ performing functions based on their fitness scores (from the combined set of current and newly mutated rewards). After $K$ iterations, we return the single reward function with the highest fitness $r^* = \arg\max_{r\in\mathcal{R}}\{f(r)\}$. This phase is detailed as function \textsc{EvoSearch} in Algorithm~\ref{alg:grace}.


\paragraph{Phase 3: Active Data Collection via Reinforcement Learning} The optimal reward $r^*$ above is obtained by inspecting existing demonstrations. To further improve the reward, we can collect further demonstrations by training a policy $\pi_{r^*}$ using the current optimal reward $r^*$; and use this policy to collect additional data $\mathcal{D}_{r^*}$.

In more detail, we employ PPO~\citep{schulman2017proximal} to train a policy in the environment of interest. As this process can be expensive, we use a predefined environment interaction budget $N$ instead of training to convergence. The new trajectories are added to the dataset of negative trajectories $D^{-}$, as they are likely to contain new edge cases the reward should consider. These are used to further refine the reward population as described in the preceding Sec.~\ref{sec:init} (Phase 2). This phase is presented as function \textsc{DataExpandRL} in Algorithm~\ref{alg:grace}.

The final algorithm, presented in Algorithm~\ref{alg:grace}, consists of repeatedly performing Evolutionary Search over reward population $\mathcal{R}$ followed by data expansion using RL-trained policy.

\paragraph{Additional reward shaping} When the reward function offline performance on $\mathcal{D}$ doesn't translate to good online RL performance, we assume that the reward signal is poorly shaped, and additional refinement is required. In these cases, the LLM's info in Eq.~\ref{eq:mutation} is augmented beyond low-fitness states to include full expert trajectories alongside the current reward value for each state. We then instruct the LLM to reshape the reward function, using expert trajectories as a reference, so that it provides a signal that increases monotonically.

\section{Experiments}
We empirically evaluate \method{} with respect to its ability to generate rewards that lead to effective policy learning. Specifically, we aim to address the following questions:
\begin{itemize}
\item \textbf{Accuracy and Generalization}: Can \method{} recover correct rewards, and with how many expert samples?
\item \textbf{Policy Learning Performance}: How does \method{} compare to other IRL methods or to online RL trained with ground-truth rewards?
\item \textbf{Qualitative Properties:} How well-shaped are the rewards produced by \method{}?
\item \textbf{Interpretability and Multi-Task Efficacy}: Does \method{} produce reusable reward APIs?
\end{itemize}

\begin{table*}[htbp]
    \centering
    \begin{tabular}{lc|ccccc}
        \toprule
         & \textbf{PPO} & \textbf{GRACE w/} & \textbf{GRACE w/} & \textbf{GAIL w/} & \textbf{GAIL w/} & \textbf{AIRL w/} \\
         & & \textbf{GPT-4o} & \textbf{Qwen3-Coder-30B} & \textbf{10 traj} & \textbf{200 traj} & \textbf{200 traj} \\
        \midrule
        \textbf{Hopper}   & $2212 \pm 54$  & $2143 \pm 80$  & $2106 \pm 76$  & $1902 \pm 183$ & $2056 \pm 92$  & $2028 \pm 82$ \\
        \addlinespace
        \textbf{Walker}   & $2675 \pm 292$ & $2072 \pm 576$ & $2229 \pm 600$ & $790 \pm 90$   & $1982 \pm 101$ & $2108 \pm 293$ \\
        \addlinespace
        \textbf{Ant}      & $6239 \pm 237$ & $5707 \pm 210$ & $6085 \pm 804$ & $3871 \pm 408$ & $5521 \pm 674$ &
        $4308 \pm 306$ \\
        \addlinespace
        \textbf{Humanoid} & $6455 \pm 302$ & $5809 \pm 106$ & $5921 \pm 301$ & $4772 \pm 251$ & $6521 \pm 337$ &
        $6512 \pm 291$ \\
        \bottomrule
    \end{tabular}
    \caption{\textbf{MuJoCo Results} Average returns on 4 MuJoCo continuous control tasks. Average and standard deviation is reported across 5 different seeds. The total number of required LLM calls to recover a reward for each task averages at 2000 for both GPT-4o and Qwen3-Coder-30B.}
    \label{tab:mujoco}
\end{table*}
\begin{figure*}
    \centering
    \includegraphics[width=0.7\textwidth]{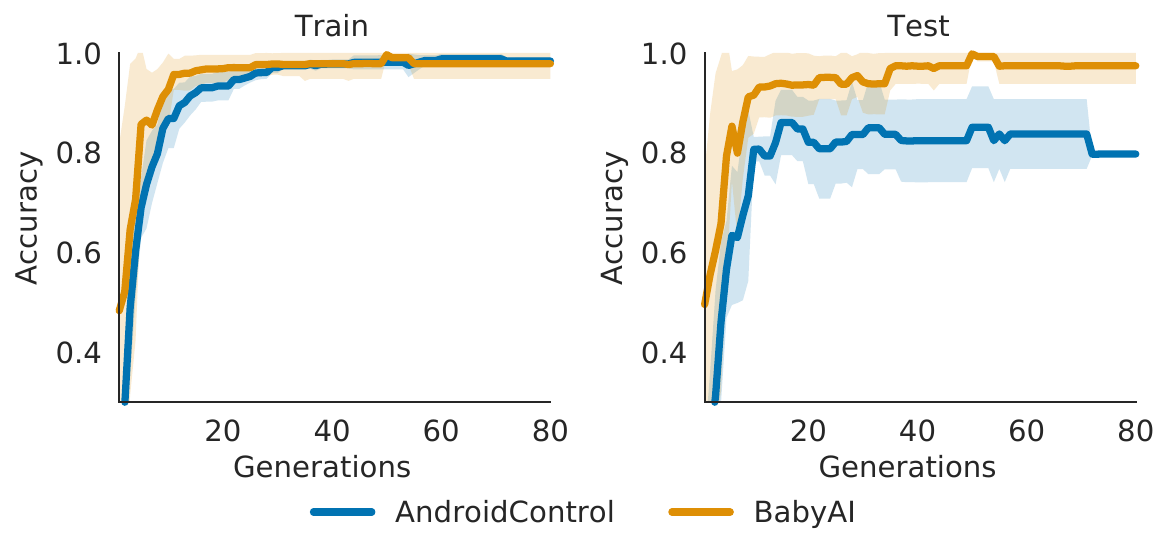}\\
    \caption{\textbf{Fitness vs Number of generations.} Evolution of train and test fitness across evolution generations, as defined by Algorithm~\ref{alg:grace}, for \emph{BabyAI} and \emph{AndroidControl} (multi-level settings). For \emph{BabyAI}, we provide 8 expert trajectories and 8 negative trajectories for each task. For \emph{AndroidControl}, we provide 8-12 expert trajectories and $\sim$800 negative trajectories total. Shading is standard deviation across 3 seeds. For these experiments, no online data is added beyond the initial trajectories provided ($M=1$).} 
    \label{fig:multi_training}
\end{figure*}
\subsection{Experimental Setup}
To evaluate \method{}, we conduct experiments in three distinct domains: the procedurally generated maze environment \emph{BabyAI}~\citep{chevalier2018babyai}, which tests reasoning and generalization; the physics simulator \emph{MuJoCo}~\citep{todorov2012mujoco}, which ensures \method{} works on continuous, non-symbolic environments; and the Android-based UI simulator \emph{AndroidWorld}~\citep{rawles2024androidworld}, to evaluate applicability to real-world, complex problems. Across all experiments, we don't provide \method{} with any in-context information about the environment (such as a description or source code) to ensure a fair comparison against baselines.

\paragraph{BabyAI} Our \emph{BabyAI} evaluation suite comprises 20 levels, including 3 custom levels designed to test zero-shot reasoning on tasks not present in public repositories, thereby mitigating concerns of data contamination. Expert demonstrations are generated using the \texttt{BabyAI-Bot}~\citep{farama2025minigrid}, which algorithmically solves \emph{BabyAI} levels optimally. We extend the bot to support our custom levels as well. For each level, we gather approximately 500 expert trajectories. Another 500 negative trajectories are collected by running a randomly initialized agent in the environment. The training dataset consists of up to 16 trajectories, including both expert and negative examples. All remaining trajectories constitute the test set. For each dataset, we evolve the reward on the train trajectories and report both train and test fitness from Eq.~(\ref{eq:fitness_grace}). For this environment, we never augment the dataset with online trajectories ($M=1$), as we establish it's not necessary to recover the optimal reward.

The state is represented by a $(h, w, 3)$ array. The state is fully observable, with the first channel containing information about the object type (with each integer corresponding to a different object, such as box, key, wall, or agent), the second channel contains information about the object's color and the third any extra information (e.g. agent direction, if is the door locked). The recovered reward function needs to operate on the state representation as a $(h, w, 3)$ array, but we also include a pixel image of the state in context for the LLM. For the classic GAIL baseline, we finetune \texttt{llava-onevision-qwen2-0.5b-ov-hf} on state images as the reward model. Given \emph{BabyAI} is an environment that terminates on success, we only consider the last state of expert trajectories as a positive state, we do this for both \method{} and the GAIL baseline.

\paragraph{MuJoCo} We conduct additional experiments on 4 tasks from the classical \emph{MuJoCo} continuous control suite~\citep{todorov2012mujoco}: \texttt{Hopper, Walker, Ant, Humanoid}. These tasks demonstrate that~\method{} also succeeds at reward design in continuous action and state spaces. We run all our MuJoCo experiments using the fully differentiable physics engine Brax~\citep{brax2021github} to speed up learning. Unlike the \emph{BabyAI} and Android experiments, in \emph{MuJoCo} we augment the dataset 10 times ($M=10$) with new trajectories coming from the learner policy. The reward is only updated if the fitness is low on the newly added trajectories.

\paragraph{Android} To assess \method{} in a high-dimensional, real-world setting, we use the AndroidControl dataset~\citep{rawles2023androidwildlargescaledataset, li2024effects}, which provides a rich collection of complex, multi-step human interactions across standard Android applications. The state space includes both raw screen pixels and the corresponding XML view hierarchy. Both representations are included in-context in the mutation prompt, while the reward function only takes the XML representation as input.

From this dataset, we curate a subset of trajectories focused on the Clock application, where users successfully complete tasks such as ``set a timer 6 hours from now''. These serve as our positive examples. Negative samples are drawn from trajectories in other applications (e.g., Calculator, Calendar, Settings). Specifically, we select 12 expert demos for the ``set timer'' task, and 8 each for the stopwatch tasks (``pause stopwatch'' and ``run stopwatch''). We use 80\% of trajectories in the train set and the remaining for the test set. For this environment, we never augment the dataset with online trajectories ($M=1$), as we establish it's not necessary to recover the optimal reward. In this setting, we recover a multi-task reward and we give the option to the LLM to remove any states it judges as unnecessary to task completion.

\textbf{\method{} Parameters} All parameters used across our experiments can be found in Appendix \ref{app:hyperparams}.

\begin{minipage}{0.52\linewidth}
\begin{table}[H]
    \centering\footnotesize
    \begin{tabular}{lc|cc}
    \toprule
    \textbf{Task} & \textbf{PPO} & \textbf{GAIL} & \textbf{\method{}} \\
    \midrule
    GoToRedBallNoDist      & 1.00 & \textbf{1.00} & \textbf{1.00} \\
    GoToRedBall            & 1.00 & 0.35 & \textbf{1.00} \\
    PickupDist             & 0.31 & 0.15 & \textbf{0.32} \\
    PickupLoc              & 0.21 & 0.00 & \textbf{0.26} \\
    GoToObj                & 1.00 & 0.92 & \textbf{1.00} \\
    OpenDoorColor          & 1.00 & 0.98 & \textbf{1.00} \\
    OpenTwoDoors           & 1.00 & 0.37 & \textbf{1.00} \\
    PlaceBetween (new)     & 0.09 & 0.01 & \textbf{0.09} \\
    OpenMatchingDoor (new) & 0.79 & 0.20 & \textbf{0.35} \\
    Multi-task             & 0.95 & 0.31 & \textbf{0.92} \\
    \bottomrule
    \end{tabular}
    \caption{\textbf{Success rates on selected BabyAI environments.} 
    \method{} compared against PPO and GAIL. 
    \method{} uses 8 expert trajectories per task, while GAIL uses 2000. Additional results are available in Table \ref{tab:table_online_rl_expanded}}
    \label{tab:table_online_rl}
\end{table}
\end{minipage}
\hfill
\begin{minipage}{0.43\linewidth}
    \begin{figure}[H]
        \centering
        \includegraphics[width=\linewidth]{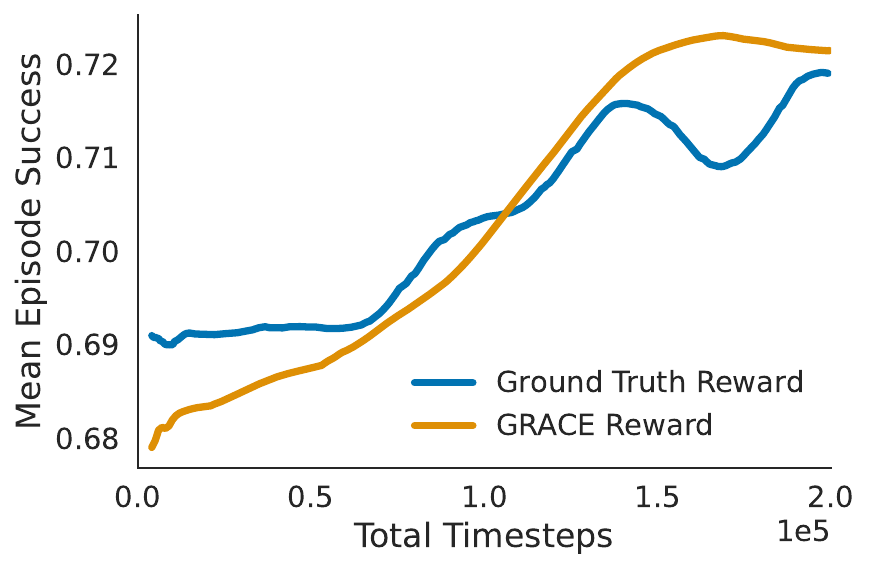}
        \caption{\textbf{Training Curves for \emph{AndroidWorld} Clock Tasks.} 
        Mean episode success over the 3 \emph{AndroidWorld} clock tasks: 
        ClockStopWatchPausedVerify, ClockStopWatchRunning, and ClockTimerEntry.}
        \label{fig:android_world}
    \end{figure}
\end{minipage}
\subsection{Analysis}
\textbf{\method{} recovers rewards with high accuracy.} We first examine whether \method{} evolutionary search (Phase 2) can successfully recover the underlying task reward from demonstrations alone. We evaluate this in two settings using \emph{BabyAI}: (i) a single-level setting, where the model infers a task-specific reward, and (ii) a more challenging multi-level setting, where \method{} must learn a single, general reward function conditioned on both state and a language goal.

\begin{figure*}[t]
    \centering
    \begin{subfigure}[b]{0.4\textwidth}
        \includegraphics[width=\textwidth]{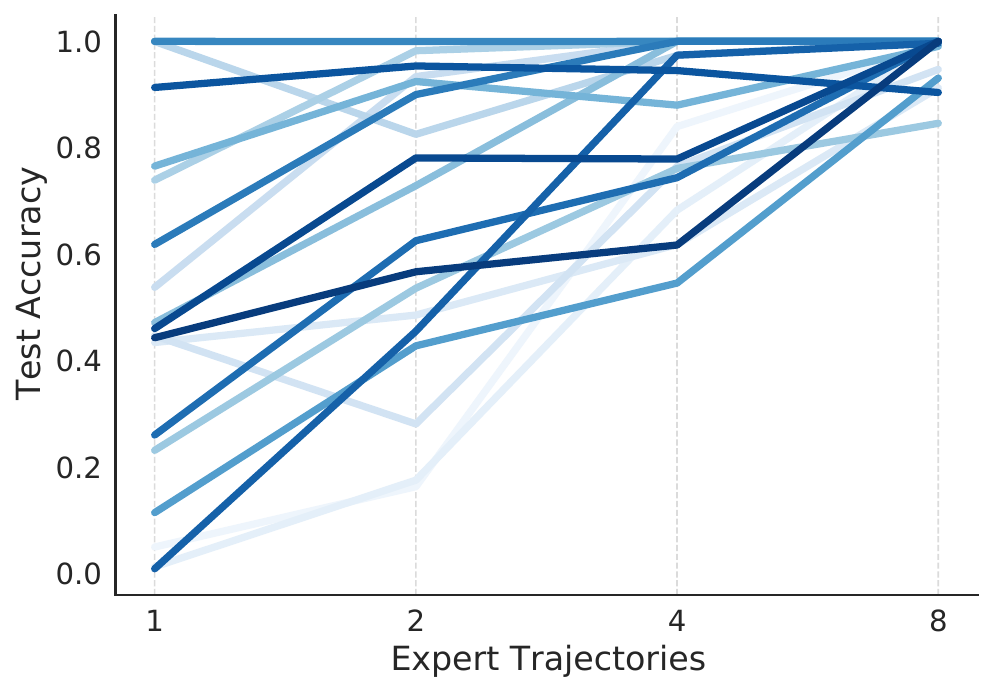}
        \caption{}
        \label{fig:num_samples_first_figure}
    \end{subfigure}
    \hfill 
    \begin{subfigure}[b]{0.4\textwidth}
        \includegraphics[width=\textwidth]{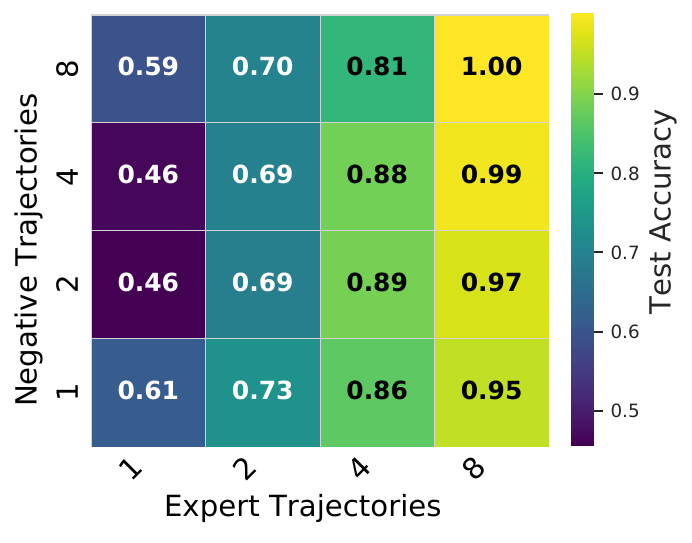}
        \caption{}
        \label{fig:num_samples_second_figure}
    \end{subfigure}
    \caption{\textbf{Fitness vs Number of Expert Trajectories.} The accuracy is computed on test dataset after obtaining maximum fitness on training data with corresponding number of expert and negative training trajectories. (a) Performance on all 20 BabyAI tasks. (b) Aggregate accuracy across 20 BabyAI tasks.}
    \label{fig:ablation_num_samples}
\end{figure*}

In Figures~\ref{fig:ablation_num_samples} and \ref{fig:multi_training}, we show that the reward accuracy consistently reaches 1.0 across all BabyAI tasks in both single- and multi-level settings, as well as on AndroidControl. A fitness of 1.0 corresponds to assigning higher values to all expert states than to negative states.

We further ablate two aspects of the algorithm. First, we analyze sample efficiency by varying the number of expert and negative demonstrations. Results on BabyAI (Figure~\ref{fig:num_samples_first_figure}) show non-trivial performance even with a single demonstration, with gradual improvement and perfect scores achieved using only eight expert trajectories. The number of negative trajectories also plays a role, though to a lesser degree: for example, accuracy of $0.95$ is achieved with just a single negative trajectory, provided that sufficient expert trajectories are available (Figure~\ref{fig:num_samples_second_figure}).

Finally, we assess the robustness and efficiency of the evolutionary process. As shown in Figure~\ref{fig:multi_training}, in the multi-task setting GRACE reliably converges to a high-fitness reward function in fewer than 100 generations (i.e., evolutionary search steps) and no additional trajectories from the learner agent ($M=1$), demonstrating the effectiveness of our LLM-driven refinement procedure.

\textbf{Comparison with Baselines}: To validate the quality of the inferred reward function, we compare against two approaches. First, we use PPO~\citep{schulman2017proximal} to optimise both the rewards recovered by \method{} as well as the groundtruth sparse reward. Clearly, the latter should serve as an oracle, while it does not benefit from dense rewards. As an IRL baseline, we compare against GAIL~\citep{ho2016gail}. GAIL is trained with a large dataset of $2,000$ expert trajectories per task.

As shown in Tables \ref{tab:mujoco} and \ref{tab:table_online_rl}, \method{} consistently matches or outperforms baselines across all tasks with less training data. On several tasks, \method{} matches Oracle PPO with ground-truth rewards, whereas GAIL completely fails. This demonstrates that the interpretable, code-based rewards from \method{} are practically effective, enabling successful downstream policy learning. To ensure a fair comparison, baseline and \method{} agents are trained using the same underlying PPO implementation, agent architecture and hyperparameters. Performance is measured by the final task success rate after 1e7 environment steps. Crucially, no extra information or environment code is provided in context to \method{}.
Similarly, we evolve a separate reward function for each task in the AndroidControl dataset matching tasks present in the Clock \emph{AndroidWorld} tasks: ClockStopWatchPausedVerify, ClockStopWatchRunning and ClockTimerEntry. The training curves for all tasks (averaged) are reported in Figure \ref{fig:android_world}.
\begin{figure*}[h]
    \centering
    \begin{subfigure}[b]{0.45\textwidth}
        \centering
        \includegraphics[width=\textwidth]{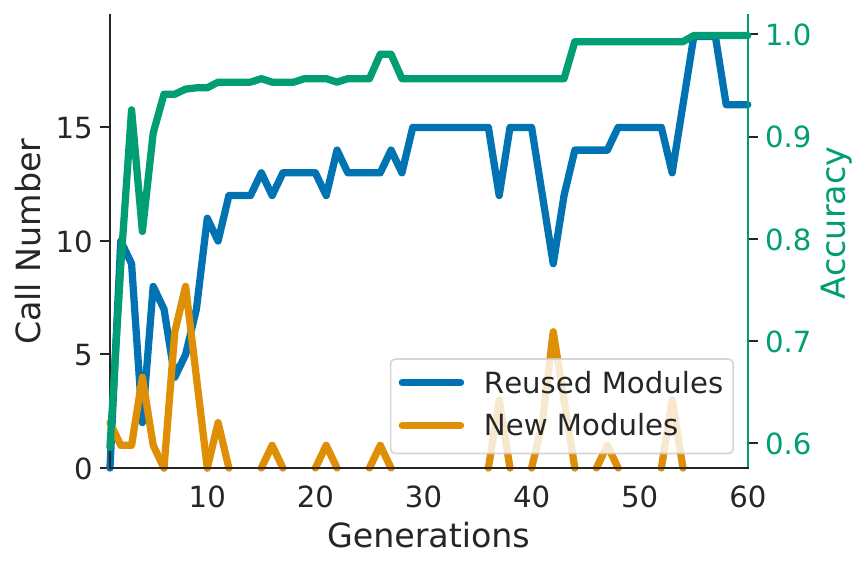}
        \label{fig:module_reuse_plot}
    \end{subfigure}
    \hfill 
    \begin{subfigure}[b]{0.45\textwidth}
        \centering
        \includegraphics[width=\textwidth]{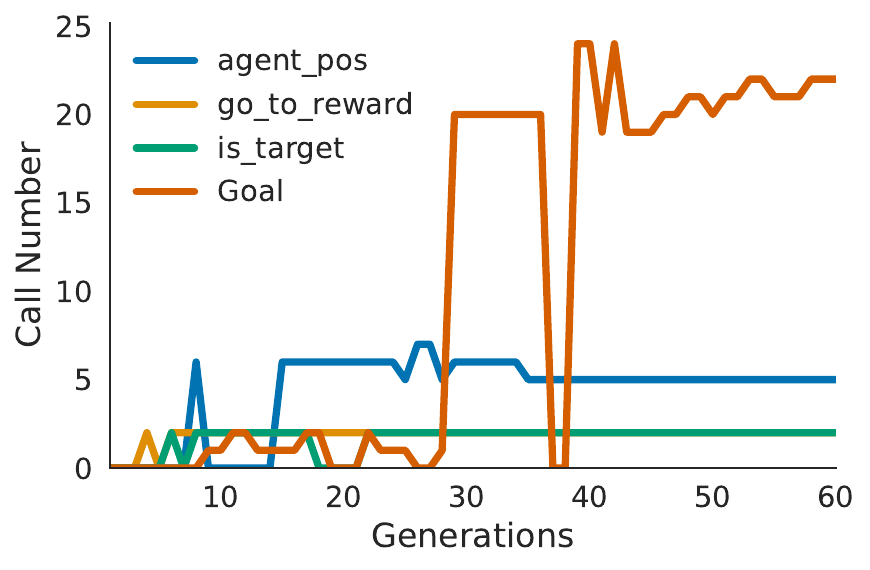}
        \label{fig:module_reuse_counts}
    \end{subfigure}
    \vspace{-0.5cm}
    \caption{\textbf{Module and function reuse across generations} On the left, we show at each generation step the number of newly created modules and the number of existing and thus reused modules from prior rewards, contrasted with the accuracy in the reward population. On the right, we show number of times a module are being re-used, for a select set of modules.} 
    \label{fig:reuse_plots}
\end{figure*}

\paragraph{MuJoCo Performance} In our MuJoCo evaluation, \method{} leverages an LLM-generated debug function to analyze distributional differences between expert and negative trajectories, computing key statistics such as mean, standard deviation, percentiles, and Cohen's $d$ for each state feature. While this statistical analysis proves effective for lower-dimensional tasks, the method still struggles with the Humanoid environment. We hypothesize that this performance drop stems from the LLM's struggle to effectively parse and reason over the extensive numerical arrays required to represent high-dimensional states in a text-based format.


\begin{figure}[t]
    \centering
    \begin{subfigure}[b]{0.45\textwidth}
        \includegraphics[width=\textwidth]{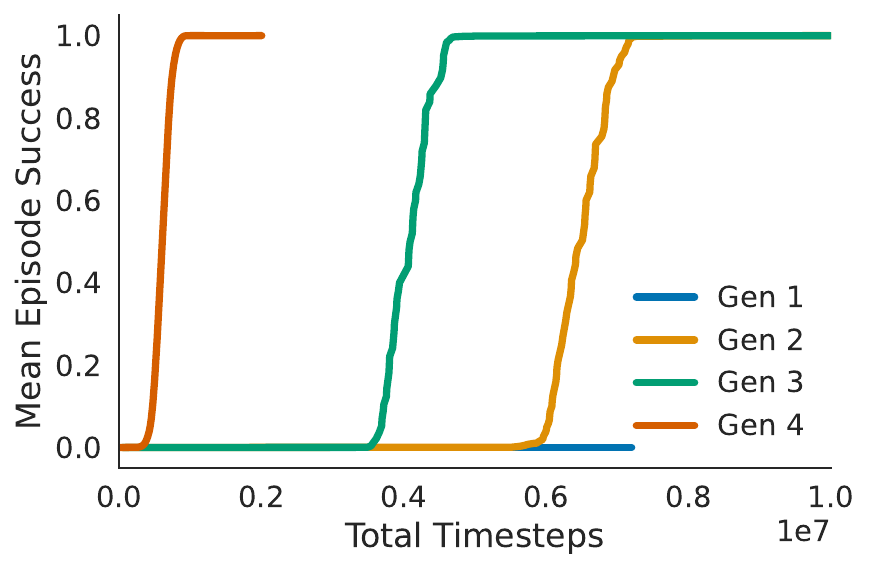}
        \label{fig:first_figure}
    \end{subfigure}
    \hfill
    \begin{subfigure}[b]{0.45\textwidth}
        \includegraphics[width=\textwidth]{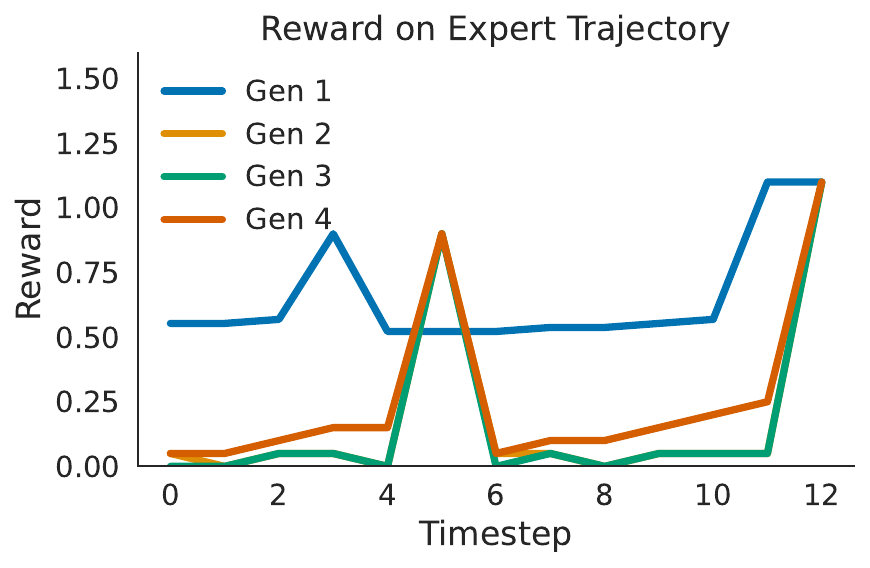}
        \label{fig:second_figure}
    \end{subfigure}
    \caption{\textbf{Shaping} Using the default reward recovered by \method{} occasionally leads to failure in learning the correct behavior due to poor shaping. Through the targeted shaping in Phase 3, we significantly improve final performance and speed of learning.}
    \label{fig:shaping}
\end{figure}

\textbf{\method{} generates well shaped rewards}: We demonstrate \method{}'s ability to produce well-shaped rewards that accelerate learning. For challenging, long-horizon tasks like OpenTwoDoors, a correct but unshaped reward can lead to local optima where the agent gets stuck (Figure \ref{fig:shaping}, ``Gen 1''). By explicitly tasking the LLM to introduce shaping terms, \method{} refines the reward to provide a denser learning signal. As shown in Figure \ref{fig:shaping}, this targeted shaping dramatically improves both the final performance and the speed of learning, allowing the agent to solve the task efficiently. This confirms that \method{} not only finds what the goal is but also learns how to guide an agent towards it.

\textbf{\method{} Code Reuse}: A key advantage of representing rewards as code is the natural emergence of reusable functions that collectively form a domain-specific reward library. We study this phenomenon in the multi-task \emph{BabyAI} setting (Figure~\ref{fig:reuse_plots}). In the early generations of evolutionary search, \method{} actively generates many new modules to explore alternative reward structures. After generation $10$, the rate of new module creation drops sharply. At this point, \method{} shifts toward reusing the most effective, high-level modules it has already discovered.

To further illustrate this reuse, Figure~\ref{fig:reuse_plots} (right) shows call counts for a selected set of modules within the evolving reward API. For instance, the \emph{Goal} module, which summarizes a set of goals, is initially used sparingly but becomes heavily invoked following a code refactor at generation $30$. Likewise, the \emph{agent\_pos} function is reused at least five times after its introduction. These trends demonstrate that \method{} progressively builds a reward library that supports efficient multi-task generalization.



\section{Discussion}
\paragraph{Limitations} A key limitation of \method{} is its limited scalability to high-dimensional state spaces for evolving reward functions. While code offers interpretability, neural networks are inherently better suited for processing high-dimensional observations (such as pixels), as they excel at learning distributed representations directly from raw sensory data, a task where symbolic feature extraction often proves brittle or infeasible.

\paragraph{Conclusion} We introduce \method{}, a novel framework that leverages LLMs within an evolutionary search to address the critical challenge of interpretability in IRL. Our empirical results demonstrate that by representing reward functions as executable code, we can move beyond the black-box models of traditional IRL and produce rewards that are transparent, verifiable, and effective in RL learning. We show that \method{} successfully recovers accurate and generalisable rewards from few expert trajectories, in stark contrast to deep IRL methods like AIRL. This sample efficiency suggests that the strong priors and reasoning capabilities of LLMs provide a powerful inductive bias. Furthermore, we demonstrate the framework's practical utility by applying it to the AndroidWorld environment, showing that \method{} can learn rewards for a variety of tasks with real-world applications directly from  user interaction data.




\bibliography{main}

@misc{sapora2024evil,
      title={EvIL: Evolution Strategies for Generalisable Imitation Learning}, 
      author={Silvia Sapora and Gokul Swamy and Chris Lu and Yee Whye Teh and Jakob Nicolaus Foerster},
      year={2024},
}

@book{puterman2014markov,
  title={Markov decision processes: discrete stochastic dynamic programming},
  author={Puterman, Martin L},
  year={2014},
  publisher={John Wiley \& Sons}
}

@article{syed2007game,
  title={A game-theoretic approach to apprenticeship learning},
  author={Syed, Umar and Schapire, Robert E},
  journal={Advances in neural information processing systems},
  volume={20},
  year={2007}
}

@inproceedings{swamy2021moments,
  title={Of moments and matching: A game-theoretic framework for closing the imitation gap},
  author={Swamy, Gokul and Choudhury, Sanjiban and Bagnell, J Andrew and Wu, Steven},
  booktitle={International Conference on Machine Learning},
  pages={10022--10032},
  year={2021},
  organization={PMLR}
}

@article{chevalier2018babyai,
  title={Babyai: A platform to study the sample efficiency of grounded language learning},
  author={Chevalier-Boisvert, Maxime and Bahdanau, Dzmitry and Lahlou, Salem and Willems, Lucas and Saharia, Chitwan and Nguyen, Thien Huu and Bengio, Yoshua},
  journal={arXiv preprint arXiv:1810.08272},
  year={2018}
}

@article{li2024effects,
  title={On the effects of data scale on computer control agents},
  author={Li, Wei and Bishop, William and Li, Alice and Rawles, Chris and Campbell-Ajala, Folawiyo and Tyamagundlu, Divya and Riva, Oriana},
  journal={arXiv e-prints},
  pages={arXiv--2406},
  year={2024}
}

@article{rawles2024androidworld,
  title={Androidworld: A dynamic benchmarking environment for autonomous agents},
  author={Rawles, Christopher and Clinckemaillie, Sarah and Chang, Yifan and Waltz, Jonathan and Lau, Gabrielle and Fair, Marybeth and Li, Alice and Bishop, William and Li, Wei and Campbell-Ajala, Folawiyo and others},
  journal={arXiv preprint arXiv:2405.14573},
  year={2024}
}

@article{zheng2023judging,
  title={Judging llm-as-a-judge with mt-bench and chatbot arena},
  author={Zheng, Lianmin and Chiang, Wei-Lin and Sheng, Ying and Zhuang, Siyuan and Wu, Zhanghao and Zhuang, Yonghao and Lin, Zi and Li, Zhuohan and Li, Dacheng and Xing, Eric and others},
  journal={Advances in neural information processing systems},
  volume={36},
  pages={46595--46623},
  year={2023}
}

@inproceedings{cookcomplexity,
author = {Cook, Stephen A.},
title = {The complexity of theorem-proving procedures},
year = {1971},
isbn = {9781450374644},
publisher = {Association for Computing Machinery},
address = {New York, NY, USA},
url = {https://doi.org/10.1145/800157.805047},
doi = {10.1145/800157.805047},
booktitle = {Proceedings of the Third Annual ACM Symposium on Theory of Computing},
pages = {151–158},
numpages = {8},
location = {Shaker Heights, Ohio, USA},
series = {STOC '71}
}

@article{klissarov2023motif,
  title={Motif: Intrinsic motivation from artificial intelligence feedback},
  author={Klissarov, Martin and D'Oro, Pierluca and Sodhani, Shagun and Raileanu, Roberta and Bacon, Pierre-Luc and Vincent, Pascal and Zhang, Amy and Henaff, Mikael},
  journal={arXiv preprint arXiv:2310.00166},
  year={2023}
}

@article{klissarov2024modeling,
  title={On the Modeling Capabilities of Large Language Models for Sequential Decision Making},
  author={Klissarov, Martin and Hjelm, Devon and Toshev, Alexander and Mazoure, Bogdan},
  journal={arXiv preprint arXiv:2410.05656},
  year={2024}
}

@article{venuto2024code,
  title={Code as reward: Empowering reinforcement learning with vlms},
  author={Venuto, David and Islam, Sami Nur and Klissarov, Martin and Precup, Doina and Yang, Sherry and Anand, Ankit},
  journal={arXiv preprint arXiv:2402.04764},
  year={2024}
}

@article{yu2023language,
  title={Language to rewards for robotic skill synthesis},
  author={Yu, Wenhao and Gileadi, Nimrod and Fu, Chuyuan and Kirmani, Sean and Lee, Kuang-Huei and Arenas, Montse Gonzalez and Chiang, Hao-Tien Lewis and Erez, Tom and Hasenclever, Leonard and Humplik, Jan and others},
  journal={arXiv preprint arXiv:2306.08647},
  year={2023}
}

@article{novikov2025alphaevolve,
  title={AlphaEvolve: A coding agent for scientific and algorithmic discovery},
  author={Novikov, Alexander and V{\~u}, Ng{\^a}n and Eisenberger, Marvin and Dupont, Emilien and Huang, Po-Sen and Wagner, Adam Zsolt and Shirobokov, Sergey and Kozlovskii, Borislav and Ruiz, Francisco JR and Mehrabian, Abbas and others},
  journal={arXiv preprint arXiv:2506.13131},
  year={2025}
}

@article{romera2024mathematical,
  title={Mathematical discoveries from program search with large language models},
  author={Romera-Paredes, Bernardino and Barekatain, Mohammadamin and Novikov, Alexander and Balog, Matej and Kumar, M Pawan and Dupont, Emilien and Ruiz, Francisco JR and Ellenberg, Jordan S and Wang, Pengming and Fawzi, Omar and others},
  journal={Nature},
  volume={625},
  number={7995},
  pages={468--475},
  year={2024},
  publisher={Nature Publishing Group UK London}
}

@article{dainese2024generating,
  title={Generating code world models with large language models guided by monte carlo tree search},
  author={Dainese, Nicola and Merler, Matteo and Alakuijala, Minttu and Marttinen, Pekka},
  journal={Advances in Neural Information Processing Systems},
  volume={37},
  pages={60429--60474},
  year={2024}
}

@article{zhou2023language,
  title={Language agent tree search unifies reasoning acting and planning in language models},
  author={Zhou, Andy and Yan, Kai and Shlapentokh-Rothman, Michal and Wang, Haohan and Wang, Yu-Xiong},
  journal={arXiv preprint arXiv:2310.04406},
  year={2023}
}

@article{madaan2023self,
  title={Self-refine: Iterative refinement with self-feedback},
  author={Madaan, Aman and Tandon, Niket and Gupta, Prakhar and Hallinan, Skyler and Gao, Luyu and Wiegreffe, Sarah and Alon, Uri and Dziri, Nouha and Prabhumoye, Shrimai and Yang, Yiming and others},
  journal={Advances in Neural Information Processing Systems},
  volume={36},
  pages={46534--46594},
  year={2023}
}

@misc{rawles2023androidwildlargescaledataset,
      title={Android in the Wild: A Large-Scale Dataset for Android Device Control}, 
      author={Christopher Rawles and Alice Li and Daniel Rodriguez and Oriana Riva and Timothy Lillicrap},
      year={2023},
      eprint={2307.10088},
      archivePrefix={arXiv},
      primaryClass={cs.LG},
      url={https://arxiv.org/abs/2307.10088}, 
}

@misc{novikov2025alphaevolvecodingagentscientific,
      title={AlphaEvolve: A coding agent for scientific and algorithmic discovery}, 
      author={Alexander Novikov and Ngân Vũ and Marvin Eisenberger and Emilien Dupont and Po-Sen Huang and Adam Zsolt Wagner and Sergey Shirobokov and Borislav Kozlovskii and Francisco J. R. Ruiz and Abbas Mehrabian and M. Pawan Kumar and Abigail See and Swarat Chaudhuri and George Holland and Alex Davies and Sebastian Nowozin and Pushmeet Kohli and Matej Balog},
      year={2025},
      eprint={2506.13131},
      archivePrefix={arXiv},
      primaryClass={cs.AI},
      url={https://arxiv.org/abs/2506.13131}, 
}

@article{schulman2017proximal,
  title={Proximal Policy Optimization Algorithms},
  author={Schulman, John and Wolski, Filip and Dhariwal, Prafulla and Radford, Alec and Klimov, Oleg},
  journal={arXiv preprint arXiv:1707.06347},
  year={2017}
}

@inproceedings{ho2016gail,
  title     = {Generative Adversarial Imitation Learning},
  author    = {Ho, Jonathan and Ermon, Stefano},
  booktitle = {Advances in Neural Information Processing Systems},
  volume    = {29},
  year      = {2016}
}

@inproceedings{ng2000irl,
  title     = {Algorithms for Inverse Reinforcement Learning},
  author    = {Ng, Andrew Y. and Russell, Stuart J.},
  booktitle = {Proceedings of the Seventeenth International Conference on Machine Learning (ICML)},
  pages     = {663--670},
  year      = {2000},
  publisher = {Morgan Kaufmann}
}

@inproceedings{abbeel2004apprenticeship,
  title     = {Apprenticeship Learning via Inverse Reinforcement Learning},
  author    = {Abbeel, Pieter and Ng, Andrew Y.},
  booktitle = {Proceedings of the Twenty-First International Conference on Machine Learning (ICML)},
  pages     = {1--8},
  year      = {2004},
  publisher = {ACM}
}

@inproceedings{ziebart2008maximum,
  title     = {Maximum Entropy Inverse Reinforcement Learning},
  author    = {Ziebart, Brian D. and Maas, Andrew L. and Bagnell, J. Andrew and Dey, Anind K.},
  booktitle = {Proceedings of the Twenty-Third AAAI Conference on Artificial Intelligence (AAAI)},
  pages     = {1433--1438},
  year      = {2008},
  publisher = {AAAI Press}
}

@inproceedings{christiano2017preferences,
  title     = {Deep Reinforcement Learning from Human Preferences},
  author    = {Christiano, Paul F. and Leike, Jan and Brown, Tom and Martic, Miljan and Legg, Shane and Amodei, Dario},
  booktitle = {Advances in Neural Information Processing Systems (NeurIPS)},
  pages     = {4299--4307},
  year      = {2017}
}

@inproceedings{wilson2007multi,
  title     = {Multi-task Reinforcement Learning: A Hierarchical Bayesian Approach},
  author    = {Wilson, Aaron and Fern, Alan and Tadepalli, Prasad},
  booktitle = {Proceedings of the 24th International Conference on Machine Learning (ICML)},
  pages     = {1015--1022},
  year      = {2007},
  publisher = {ACM}
}

@inproceedings{teh2017distral,
  title     = {Distral: Robust Multitask Reinforcement Learning},
  author    = {Teh, Yee Whye and Bapst, Victor and Czarnecki, Wojciech M. and Quan, John and Kirkpatrick, James and Hadsell, Raia and Heess, Nicolas and Pascanu, Razvan},
  booktitle = {Advances in Neural Information Processing Systems (NeurIPS)},
  pages     = {4499--4509},
  year      = {2017}
}

@inproceedings{parisotto2015actor,
  title     = {Actor-Mimic: Deep Multitask and Transfer Reinforcement Learning},
  author    = {Parisotto, Emilio and Ba, Jimmy Lei and Salakhutdinov, Ruslan},
  booktitle = {International Conference on Learning Representations (ICLR)},
  year      = {2016}
}

@InProceedings{pmlr-v235-venuto24a,
  title = 	 {Code as Reward: Empowering Reinforcement Learning with {VLM}s},
  author =       {Venuto, David and Islam, Mohammad Sami Nur and Klissarov, Martin and Precup, Doina and Yang, Sherry and Anand, Ankit},
  booktitle = 	 {Proceedings of the 41st International Conference on Machine Learning},
  pages = 	 {49368--49387},
  year = 	 {2024},
  editor = 	 {Salakhutdinov, Ruslan and Kolter, Zico and Heller, Katherine and Weller, Adrian and Oliver, Nuria and Scarlett, Jonathan and Berkenkamp, Felix},
  volume = 	 {235},
  series = 	 {Proceedings of Machine Learning Research},
  month = 	 {21--27 Jul},
  publisher =    {PMLR},
  url = 	 {https://proceedings.mlr.press/v235/venuto24a.html},
}

@article{ma2023eureka,
    title   = {Eureka: Human-Level Reward Design via Coding Large Language Models},
    author  = {Yecheng Jason Ma and William Liang and Guanzhi Wang and De-An Huang and Osbert Bastani and Dinesh Jayaraman and Yuke Zhu and Linxi Fan and Anima Anandkumar},
    year    = {2023},
    journal = {arXiv preprint arXiv: Arxiv-2310.12931}
}

@book{goldberg1989geneticalgorithms,
  title     = {Genetic Algorithms in Search, Optimization, and Machine Learning},
  author    = {Goldberg, David E.},
  year      = {1989},
  publisher = {Addison-Wesley}
}

@book{eiben2003evolutionary,
  title     = {Introduction to Evolutionary Computing},
  author    = {Eiben, Agoston E. and Smith, James E.},
  year      = {2003},
  publisher = {Springer}
}

@inproceedings{salimans2017es,
  title     = {Evolution Strategies as a Scalable Alternative to Reinforcement Learning},
  author    = {Salimans, Tim and Ho, Jonathan and Chen, Xi and Sidor, Szymon and Sutskever, Ilya},
  booktitle = {Proceedings of the 36th International Conference on Machine Learning (ICML)},
  year      = {2017}
}

@article{romera2024funsearch,
  title   = {Mathematical discoveries from program search with large language models},
  author  = {Romera-Paredes, Bernardino and Barekatain, Mohammadamin and Novikov, Alexander and Balog, Matej and Kumar, M. Pawan and Dupont, Emilien and Ruiz, Francisco J. R. and Ellenberg, Jordan S. and Wang, Pengming and Fawzi, Omar and Kohli, Pushmeet and Fawzi, Alhussein},
  journal = {Nature},
  volume  = {625},
  number  = {7995},
  pages   = {468--475},
  year    = {2024},
  doi     = {10.1038/s41586-023-06924-6}
}

@inproceedings{sutton1999policy,
  title     = {Policy Gradient Methods for Reinforcement Learning with Function Approximation},
  author    = {Sutton, Richard S. and McAllester, David and Singh, Satinder and Mansour, Yishay},
  booktitle = {Advances in Neural Information Processing Systems (NeurIPS)},
  volume    = {12},
  pages     = {1057--1063},
  year      = {1999}
}

@inproceedings{ratliff2006maximum,
  title={Maximum Margin Planning},
  author={Ratliff, Nathan and Bagnell, J. Andrew and Zinkevich, Martin},
  booktitle={Proceedings of the 23rd International Conference on Machine Learning (ICML)},
  pages={729--736},
  year={2006},
  organization={ACM}
}

@book{molnar2020interpretable,
  title={Interpretable machine learning},
  author={Molnar, Christoph},
  year={2020},
  publisher={Lulu. com}
}

@inproceedings{wang2023llm,
    title={LLM-Blender: Ensembling Large Language Models with Pairwise Ranking and Generation},
    author={Wang, Yuxiang and Lin, Yuchen and Jiang, Dongfu and Chen, Bill Y. and Shen, Xiang and Zhao, Jidong and Yu, Xiang and Li, Chen and Qin, Xiao and Sun, Jie},
    booktitle={Proceedings of the 61st Annual Meeting of the Association for Computational Linguistics (Volume 1: Long Papers)},
    year={2023},
    publisher={Association for Computational Linguistics}
}

@article{zankner2024critique,
      title={Critique-out-Loud Reward Models}, 
      author={William Zankner and Rohan Mehta and Eric Wallace and Jack Fitzsimons and Y. Yang and Alex Mei and Daniel Levy and William S. Moses and Joseph E. Gonzalez},
      year={2024},
      eprint={2408.11791},
      archivePrefix={arXiv},
      primaryClass={cs.CL},
      url={https://arxiv.org/abs/2408.11791},
}

@letter{godel,
      title={Letter to John Von Neumann}, 
      author={Kurt Godel},
      year={1956},
      url={https://ecommons.cornell.edu/server/api/core/bitstreams/46aef9c4-288b-457d-ab3e-bb6cb1a4b88e/content},
}

@misc{salimans2017evolutionstrategiesscalablealternative,
      title={Evolution Strategies as a Scalable Alternative to Reinforcement Learning}, 
      author={Tim Salimans and Jonathan Ho and Xi Chen and Szymon Sidor and Ilya Sutskever},
      year={2017},
      eprint={1703.03864},
      archivePrefix={arXiv},
      primaryClass={stat.ML},
      url={https://arxiv.org/abs/1703.03864}, 
}

@book{evolutionarycomputing,
author = {Eiben, Agoston E. and Smith, J. E.},
title = {Introduction to Evolutionary Computing},
year = {2003},
isbn = {3540401849},
publisher = {SpringerVerlag}
}

@article{ouyang2022training,
  title={Training language models to follow instructions with human feedback},
  author={Ouyang, Long and Wu, Jeff and Jiang, Xu and Almeida, Diogo and Wainwright, Carroll and Mishkin, Pamela and Zhang, Chong and Agarwal, Sandhini and Slama, Katarina and Ray, Alex and Schulman, John and Hilton, Jacob and Kelton, Fraser and Miller, Luke and Simens, Maddie and Askell, Amanda and Welinder, Peter and Christiano, Paul and Leike, Jan and Lowe, Ryan},
  journal={arXiv preprint arXiv:2203.02155},
  year={2022}
}

@book{geneticalgorithmsearch,
author = {Goldberg, David E.},
title = {Genetic Algorithms in Search, Optimization and Machine Learning},
year = {1989},
isbn = {0201157675},
publisher = {Addison-Wesley Longman Publishing Co., Inc.},
address = {USA},
edition = {1st}
}

@article{chen2021codex,
  title     = {Evaluating Large Language Models Trained on Code},
  author    = {Chen, Mark and Tworek, Jerry and Jun, Heewoo and Yuan, Qiming and de Oliveira Pinto, Henrique Ponde and Kaplan, Jared and Edwards, Harri and Burda, Yuri and Joseph, Nicholas and Brockman, Greg and Ray, Alex and Puri, Raul and Krueger, Gretchen and Petrov, Michael and Khlaaf, Heidy and Sastry, Girish and Mishkin, Pamela and Chan, Brooke and Gray, Scott and Ryder, Nick and Pavlov, Michael and Power, Alethea and Kaiser, Lukasz and Bavarian, Miljan and Winter, Clemens and Tillet, Phil and Such, Felipe and Cummings, Dave and Plappert, Matthias and Chantzis, Fotios and Barnes, Elizabeth and Herbert-Voss, Ariel and Guss, William and Nichol, Alex and Paino, Igor and Tezak, Nikolas and Tang, Jie and Babuschkin, Igor and Balaji, Suchir and Jain, Suyog and Saunders, William and Hesse, Christopher and Carr, Mark and Lewkowycz, Aitor and Dohan, David and Mao, Howard and Thompson, Lily and Frank, Erica and Chen, Joshua and Butoi, Victor and Hernandez, David and DasSarma, Liane and Chan, Maxwell and Litwin, Mateusz and Gray, Scott and Clark, Jack and Berner, Christopher and McCandlish, Sam and Radford, Alec and Sutskever, Ilya and Amodei, Dario},
  journal   = {arXiv preprint arXiv:2107.03374},
  year      = {2021}
}

@article{austin2021programsynthesis,
  title     = {Program Synthesis with Large Language Models},
  author    = {Austin, Jacob and Odena, Augustus and Nye, Maxwell and Bosma, Maarten and Michalewski, Henryk and Dohan, David and Jiang, Ellen and Cai, Carrie and Terry, Michael and Le, Quoc V. and Sutton, Charles},
  journal   = {arXiv preprint arXiv:2108.07732},
  year      = {2021}
}

@article{li2023starcoder,
  title     = {StarCoder: may the source be with you!},
  author    = {Li, Raymond and Allal, Loubna Ben and Zi, Yacine Jernite and Kocetkov, Denis and Mou, Chenxi and Piktus, Aleksandra and Weber, Laura and Xiao, Wenhao and Bibi, Jihad and Biderman, Stella and others},
  journal   = {arXiv preprint arXiv:2305.06161},
  year      = {2023}
}

@inproceedings{fried2022incoder,
  title     = {InCoder: A Generative Model for Code Infilling and Synthesis},
  author    = {Fried, Daniel and Ainslie, Joshua and Grangier, David and Linzen, Tal and Yogatama, Dani},
  booktitle = {International Conference on Learning Representations (ICLR)},
  year      = {2022}
}

@inproceedings{nijkamp2022codegen,
  title     = {CodeGen: An Open Large Language Model for Code with Multi-Turn Program Synthesis},
  author    = {Nijkamp, Erik and Pang, Richard and Hayashi, Hiroaki and He, Tian and Roziere, Baptiste and Xu, Canwen and Li, Susan and Jurafsky, Dan and Zettlemoyer, Luke and Stoyanov, Veselin and Chung, Hyung Won},
  booktitle = {International Conference on Learning Representations (ICLR)},
  year      = {2022}
}

@inproceedings{zala2024envgen,
  title={EnvGen: Generating and Adapting Environments via LLMs for Training Embodied Agents},
  author={Zala, Abhay and Cho, Jaemin and Lin, Han and Yoon, Jaehong and Bansal, Mohit},
  booktitle={Conference on Language Modeling (CoLM)},
  year={2024}
}

@inproceedings{faldor2025omni-epic,
  title={OMNI-EPIC: Open-endedness via Models of human Notions of Interestingness with Environments Programmed in Code},
  author={Faldor, Maxence and Zhang, Jenny and Cully, Antoine and Clune, Jeff},
  booktitle={International Conference on Learning Representations},
  year={2025},
  url={https://openreview.net/forum?id=Y1XkzMJpPd}
}

@misc{farama2025minigrid,
  title        = {Minigrid: Modular \& Customizable Reinforcement Learning Environments},
  author       = {{Farama Foundation} and Chevalier-Boisvert, Maxime and Dai, Bolun and Towers, Mark and Perez-Vicente, Rodrigo and Willems, Lucas and Lahlou, Salem and Pal, Suman and Castro, Pablo Samuel and Terry, Jordan},
  year         = {2025},
  howpublished = {\url{https://github.com/Farama-Foundation/Minigrid}},
  note         = {Accessed: 2025-09-24}
}

@inproceedings{todorov2012mujoco,
  title={MuJoCo: A physics engine for model-based control},
  author={Todorov, Emanuel and Erez, Tom and Tassa, Yuval},
  booktitle={2012 IEEE/RSJ International Conference on Intelligent Robots and Systems},
  pages={5026--5033},
  year={2012},
  organization={IEEE},
  doi={10.1109/IROS.2012.6386109}
}

@software{brax2021github,
  author = {C. Daniel Freeman and Erik Frey and Anton Raichuk and Sertan Girgin and Igor Mordatch and Olivier Bachem},
  title = {Brax - A Differentiable Physics Engine for Large Scale Rigid Body Simulation},
  url = {http://github.com/google/brax},
  version = {0.14.0},
  year = {2021},
}

@misc{fu2018learningrobustrewardsadversarial,
      title={Learning Robust Rewards with Adversarial Inverse Reinforcement Learning}, 
      author={Justin Fu and Katie Luo and Sergey Levine},
      year={2018},
      eprint={1710.11248},
      archivePrefix={arXiv},
      primaryClass={cs.LG},
      url={https://arxiv.org/abs/1710.11248}, 
}
\bibliographystyle{iclr2026_conference}

\appendix
\section{Appendix}
\label{sec:appendix}

\subsection{Additional Online Results}
\begin{table}[htbp]
\centering
\begin{tabular}{lccc}
\toprule
\textbf{Task} & \textbf{PPO} & \textbf{GAIL} & \textbf{\method{}} \\
 &  & w/ 2000 trajs & w/ 8 trajs \\
\midrule
GoToRedBallNoDist      & 1.00  & 1.00 & 1.00 \\
GoToRedBall            & 1.00  & 0.35  & 1.00 \\
PickupDist             & 0.31  & 0.15  & 0.32 \\
PickupLoc              & 0.21  & 0.00  & 0.26 \\
GoToObj                & 1.00  & 0.92  & 1.00 \\
OpenDoorColor          & 1.00  & 0.98  & 1.00 \\
OpenTwoDoors           & 1.00  & 0.37  & 1.00 \\
OpenRedDoor            & 1.00  & 1.00  & 1.00 \\
GoToObjS4              & 1.00  & 1.00  & 1.00 \\
GoToRedBlueBall        & 0.96  & 0.40  & 0.99 \\
GoToRedBallGrey        & 0.97  & 0.77  & 0.99 \\
Pickup                 & 0.10  & 0.00  & 0.09 \\
Open                   & 0.30  & 0.18  & 0.22 \\
OpenRedBlueDoors       & 1.00  & 0.96  & 0.98 \\
OpenDoorLoc            & 0.39  & 0.40  & 1.00 \\
GoToLocalS8N7          & 0.64  & 0.39  & 0.97 \\
GoToDoor               & 0.74  & 0.37  & 0.99 \\
SortColors (new)       & 0.00  & 0.00  & 0.00 \\
PlaceBetween (new)     & 0.09  & 0.01  & 0.09 \\
OpenMatchingDoor (new) & 0.79  & 0.20  & 0.35 \\
Multi-task             & 0.95  & 0.31  & 0.92     \\
\bottomrule
\end{tabular}
\caption{\textbf{Success rates on additional BabyAI environments}. The performance of our method, \method{}, is compared against two key baselines: PPO, trained on the ground-truth reward, and GAIL, trained using 2000 expert trajectories per task. \method{}'s performance is evaluated with 8 expert trajectories per task to demonstrate its high sample efficiency. All values represent the final success rate at the end of training. We don't report GAIL's performance on 8 expert trajectories as it is near 0 for most tasks.}
\label{tab:table_online_rl_expanded}
\end{table}

\subsection{Extended Discussion and Future Work}
\method{}'s reliance on programmatic reward functions introduces several limitations, particularly when compared to traditional deep neural network based approaches. These limitations also point toward promising directions for future research.

\paragraph{Input modality} While generating rewards as code offers interpretability and sample efficiency, it struggles in domains where the reward depends on complex, high-dimensional perceptual inputs. Code is inherently symbolic and structured, making it less suited for interpreting raw sensory data like images or audio. For instance, creating a programmatic reward for a task like ``navigate to the cat'' is non-trivial, as ``cat'' is a difficult visual concept. NNs, in contrast, excel at learning features directly from this kind of data. Programmatic rewards can also be brittle: a small difference in the environment violating a hard-coded assumption could cause the reward logic to fail completely, whereas NNs often degrade more gracefully.

\paragraph{Data Quantity} \method{} demonstrates remarkable performance with very few demonstrations. This is a strength in data-scarce scenarios. However, it is a limitation when vast amounts of data are available. Deep IRL methods like GAIL are designed to scale with data and may uncover complex patterns from millions of demonstrations that would be difficult to capture in an explicit program. While GRACE's evolutionary search benefits from tight feedback on a small dataset, it is not clear how effectively it could leverage a massive, complex dataset.

\paragraph{Failure Cases} Although \method{} is highly sample-efficient, it can still fail. For example, in the BabyAI-OpenTwoDoors task, \method{} often proposed a reward that didn't take into account the order in which the doors were being opened. Similarly, in the new BabyAI-SortColors task, it would sometimes return a reward that only accounted for picking up and dropping both objects, without paying attention to where they were being dropped. While these errors can be easily fixed by providing a relevant negative trajectory or by treating all learner-generated states as negative trajectories, they highlight that \method{} can still misinterpret an agent's true intent based on expert demonstrations alone.

\paragraph{Hybrid Approaches} These limitations can be substantially mitigated by extending the \method{} framework to incorporate tool use, combining the strengths of both systems. The LLM could be granted access to a library of pre-trained models (e.g., object detectors, image classifiers, or segmentation models). The LLM's task would then shift from writing low-level image processing code to writing high-level logic that calls these tools and reasons over their outputs. A final direction involves generating hybrid reward functions that are part code and part neural network. The LLM could define the overall structure, logic, and shaping bonuses in code, but instantiate a small, learnable NN module for a specific, difficult-to-program component of the reward. This module could then be fine-tuned using the available demonstrations, creating a reward function that is both largely interpretable and capable of handling perceptual nuance. By exploring these hybrid approaches, future iterations of \method{} could retain the benefits of interpretability and sample efficiency while overcoming the inherent limitations of purely programmatic solutions in complex, perception-rich environments.



\newpage
\newpage
\subsection{New BabyAI Levels}
To evaluate the generalization and reasoning capabilities of \method{} and mitigate concerns of data contamination from pre-existing benchmarks, we designed three novel BabyAI levels.

\paragraph{PlaceBetween}
The agent is placed in a single room with three distinct objects (e.g., a red ball, a green ball, and a blue ball). The instruction requires the agent to pick up a specific target object and place it on an empty cell that is strictly between the other two anchor objects. Success requires being on the same row or column as the two anchors, creating a straight line. This task moves beyond simple navigation, demanding that the agent understand the spatial relationship ``between'' and act upon a configuration of three separate entities.

\paragraph{OpenMatchingDoor}
This level is designed to test indirect object identification and chained inference. The environment consists of a single room containing one key and multiple doors of different colors. The instruction is to ``open the door matching the key''. The agent cannot solve the task by simply parsing an object and color from the instruction. Instead, it must first locate the key, visually identify its color, and then find and open the door of the corresponding color. This task assesses the agent's ability to perform a simple chain of reasoning: find object A, infer a property from it, and then use that property to identify and interact with target object B.

\paragraph{SortColors}
The environment consists of two rooms connected by a door, with a red ball in one room and a blue ball in the other. The instruction is a compound goal: ``put the red ball in the right room and put the blue ball in the left room''. To make the task non-trivial, the objects' initial positions are swapped relative to their goal locations. The agent must therefore execute a sequence of sub-tasks for each object: pick up the object, navigate to the other room, and drop it. This level tests the ability to decompose a complex language command and carry out a plan to satisfy multiple, distinct objectives.
\newpage
\subsection{Hyperparameters}
\label{app:hyperparams}
\begin{table}[hbt!]
\centering
\caption{Hyperparameters for Training BabyAI with PPO}
\label{tab:hyperparams_babyai}
\begin{tabular}{l|l}
Parameter & Value \\ \hline
Base Model & llava-onevision-qwen2-0.5b-ov-hf \\
Gamma & 0.999 \\
Learning Rate & 3e-5 \\
Entropy Coef & 1e-5 \\
Num Envs & 10 \\
Num Steps & 64 \\
Episode Length & 100 \\
PPO Epochs & 2 \\
Num Minibatch & 6 \\
\end{tabular}
\end{table}

\begin{table}[hbt!]
\centering
\caption{Hyperparameters for Training AndroidWorld}
\label{tab:hyperparams_aw}
\begin{tabular}{l|l}
Parameter & Value \\ \hline
Base Model & Qwen2.5-VL-3B-Instruct \\
LoRA Rank & 512 \\
LoRA Alpha & 32 \\
LoRA Dropout & 0.1 \\
Critic Hidden Size & 2048 \\
Critic Depth & 4 \\
Gamma & 0.999 \\
Learning Rate & 3e-5 \\
Entropy Coef & 0.0 \\
Num Envs & 16 \\
Num Steps & 16 \\
Episode Length & 20 \\
PPO Epochs & 2 \\
Num Minibatch & 2 \\
\end{tabular}
\end{table}

\begin{table}[hbt!]
\centering
\caption{Hyperparameters for \method{} Evolution}
\label{tab:hyperparams_grace}
\begin{tabular}{l|l}
Parameter & Value \\ \hline
Population Size & 20 \\
Elite & 4 \\
Num Generations & 100 \\
Include expert trajectory chance & 0.25 \\
Incorrect state only chance & 0.5 \\
Expert state only chance &  0.75 \\
Model &  gpt-4o \\
\end{tabular}
\end{table}
\newpage

\subsection{Evolution Examples}
\begin{center} 
\begin{lstlisting}[style=python-diff]
def _parse_colour_from_text(text: Optional[str]) -> Optional[int]:
    if text is None:
        return None

    colour_words: Dict[str, int] = {
        "red": 0,
        "green": 1,
        "blue": 2,
|\del{"yellow": 3,}|        "purple": |\add{3,}|
|\add{        "yellow":}| 4,
        "orange": 5,  |\add{\# keep old mapping}|
|\add{        "grey": 5,  \# alias for the observed colour code in the trajectory}|
|\add{        "gray": 5,}|
    }
    |\add{lower = text.lower()}|
    for word, code in colour_words.items():
        if word in |\del{text.lower():}||\add{lower:}|
            return code
    return None


def _parse_goal_type(text: Optional[str]) -> str:
    if text is None:
        return "key"
    txt = text.lower()
    if "ball" in txt:
        return "ball"
    |\add{if "box" in txt:}|
|\add{        return "box"}|
    return "key"
\end{lstlisting}
    \captionof{figure}{\textbf{\method{} iteratively refines the initial BabyAI reward function (\color{red}iteration 0\color{black}) to handle unseen entities (\color{green}iteration 10\color{black}).} Using execution traces, the agent fixes its color code mistake and adds a new \texttt{box} entity.}
    \label{fig:babyai_codediff_short}
\end{center}

\begin{center} 
\begin{lstlisting}[style=python-diff]
from __future__ import annotations

import re
from typing import Optional, Tuple

import numpy as np

COLOR2ID = {
    "red": 0,
    "green": 1,
    "blue": 2,
    "purple": 3,
    "yellow": 4,
    "grey": 5,
    "gray": 5,  # US spelling
}

OBJECT2ID = {
    "empty": 0,
    "wall": 1,
    "floor": 2,
    "door": 3,
    "key": 5,
    "ball": 6,
    "box": 8,
    "agent": 10,
}

|\add{DIR2VEC: dict[int, Tuple[int, int]]}| ={
|\add{    0: (1, 0),  \# south}|
|\add{    1: (0, 1),  \# east}|
|\add{    2: (-1, 0),  \# north}|
|\add{    3: (0, -1),  \# west}|
}

def _parse_goal(extra_info: str ) -> Tuple[int, Optional[int]]:
|\add{    """Return *(object\_id, colour\_id)* parsed from *extra\_info*."""}|
    if not extra\_info:
        raise ValueError("extra_info must specify the target, e.g. 'the red ball'.")

    tokens = re.findall(r"[a-zA-Z]+", extra\_info.lower())
    obj_id: Optional[int] = None
    col_id: Optional[int] = None
    for tok in tokens:
        |\del{if obj\_id is None and tok in OBJECT2ID:}|
        |\add{if tok in COLOR2ID and col\_id is None:}|
            |\add{col\_id = COLOR2ID[tok]}|
        |\add{if tok in OBJECT2ID and obj\_id is None:}|
            obj_id = OBJECT2ID[tok]
        |\del{if col\_id is None and tok in COLOR2ID:}|
            |\del{col\_id = COLOR2ID[tok]}|
    if obj_id is None:
        raise ValueError(
            f"Could not parse target object from extra_info='{extra_info}'."
        )
    return obj_id, col_id  # colour may be None (wild-card)
        

class Reward:
    """Success |\del{when}||\add{definition (single-step, dense reward):}|
|\add{        100.0 – the **first** cell in front of}| the agent |\del{is *either*}|
|\del{        – on / adjacent}||\add{(according}| to the
                 |\del{closest target object (Manhattan distance <= 1), OR}|
|\del{        –}||\add{direction stored in}| the |\add{third observation channel) contains a}|
|\add{                 matching}| target |\del{has disappeared from the observable grid (picked up).}|

|\del{    Shaping:  r = 1 / (1+d) with d the Manhattan distance to the closest}|
|\del{    still-visible target, clipped at 0}||\add{object.}|
|\add{          <1.0 – shaping reward 1/(d+1) otherwise.}|
|\add{          0.0  –}| if |\add{either}| the |\add{agent or (a matching)}| target is |\del{out of view.}||\add{not visible.}|

|\add{    The implementation is modular so new goal}|
|\add{    types can be handled by extending the OBJECT/COLOR lookup tables or by}|
|\add{    replacing the *success predicate*.}|
    """

    |\del{SUCCESS\_REWARD = 100.0}|
    def __init__(self, extra\_info: |\del{Optional[str]}||\add{str}| = None):
        |\del{self.tgt\_obj\_id, self.tgt\_col\_id}||\add{self.\_target\_obj\_id, self.\_target\_colour\_id}| = _parse_goal(extra_info)

    def|\del{\_\_call\_\_(self, state: np.ndarray) -> float:  \# enable direct call}|
|\del{        return self.reward\_fn(state)}|

    def reward_fn(self, state: np.ndarray) -> float:
        |\add{"""state: (H, W, 3)}"""|
        agent_pos = self._find_agent(state)
        if agent_pos is None:
            return 0.0

        |\del{\# mask of all target objects still visible}|
        |\del{tgt\_mask = (state[:, :, 0] == self.tgt\_obj\_id) \& (}|
        |\del{    state[:, :, 1] == self.tgt\_col\_id}|
        |\del{)}|
        |\del{if not tgt\_mask.any():}|
        |\del{    \# object gone -> picked up / carried}|
        |\del{    return self.SUCCESS\_REWARD}|
        |\del{\# distance to the closest visible target}|
        |\del{tgt\_positions = np.argwhere(tgt\_mask)}|
        |\del{dists = np.abs(tgt\_positions - agent\_pos).sum(axis=1)}|
        |\add{target\_positions = self.\_find\_targets(state)}|
        |\add{if target\_positions.size == 0:}|
        |\add{    \# No matching target in view -> no shaping.}|
        |\add{    return 0.0}|
        |\add{\# ------------------------------------------------------------------}|
        |\add{\# Success predicate – target must be directly in front of the agent.}|
        |\add{\# ------------------------------------------------------------------}|
        |\add{if self.\_is\_target\_in\_front(agent\_pos, state):}|
        |\add{    return 100.0}|
        |\add{\# ------------------------------------------------------------------}|
        |\add{\# Shaping: inverse Manhattan distance (< 1.0) to the *nearest* target.}|
        |\add{\# ------------------------------------------------------------------}|
        |\add{dists = np.abs(target\_positions - agent\_pos).sum(axis=1)}|
        min_dist = int(dists.min())
        |\del{if min\_dist <= 1:}|
        |\del{    return self.SUCCESS\_REWARD}|
        return 1.0 / (1.0 + min_dist)

    @staticmethod
    def _find_agent(state: np.ndarray) -> Optional[np.ndarray]:
        |\del{"""Return (row, col) of}||\add{"""Locate}| the|\del{first}| agent |\del{pixel found,}||\add{in the observation (row, col)}| or |\del{None."""}||\add{*None* if absent."""}|
        locs = np.argwhere(state[:, :, 0] == OBJECT2ID["agent"])
        if locs.size == 0:
            return None
        return locs[0]

    |\add{def \_find\_targets(self, state: np.ndarray) -> np.ndarray:}|
|\add{        """Return an (N, 2) array of row/col positions of matching targets."""}|
|\add{        obj\_mask = state[:, :, 0] == self.\_target\_obj\_id}|
|\add{        if self.\_target\_colour\_id is not None:}|
|\add{            col\_mask = state[:, :, 1] == self.\_target\_colour\_id}|
|\add{            mask = obj\_mask \& col\_mask}|
|\add{        else:}|
|\add{            mask = obj\_mask}|
|\add{        return np.argwhere(mask)}|

|\add{    def \_is\_target\_in\_front(self, agent\_pos: np.ndarray, state: np.ndarray) -> bool:}|
|\add{        """Return *True* iff the cell directly in front of the agent matches target."""}|
|\add{        row, col = agent\_pos}|
|\add{        agent\_dir = int(state[row, col, 2])}|
|\add{        drow, dcol = DIR2VEC.get(agent\_dir, (1, 0))  \# default to south if unknown}|
|\add{        f\_row, f\_col = row + drow, col + dcol}|

|\add{        \# Out of bounds → cannot be success.}|
|\add{        if not (0 <= f\_row < state.shape[0] and 0 <= f\_col < state.shape[1]):}|
|\add{            return False}|

|\add{        \# Check object id}|
|\add{        if state[f\_row, f\_col, 0] != self.\_target\_obj\_id:}|
|\add{            return False}|

|\add{        \# Check colour if colour was specified.}|
|\add{        if (}|
|\add{            self.\_target\_colour\_id is not None}|
|\add{            and state[f\_row, f\_col, 1] != self.\_target\_colour\_id}|
|\add{        ):}|
|\add{            return False}|

|\add{        return True}|
\end{lstlisting}
    \captionof{figure}{Example of code evolution across many generations.}
    \label{fig:babyai_codediff_long_1}
\end{center}

\newpage
\subsection{Generated Rewards}
\label{app:gen_rewards}
\definecolor{codegreen}{rgb}{0,0.6,0}
\definecolor{codegray}{rgb}{0.5,0.5,0.5}
\definecolor{codepurple}{rgb}{0.58,0,0.82}
\definecolor{backcolour}{rgb}{0.95,0.95,0.92}

\lstdefinestyle{mystyle}{
    backgroundcolor=\color{backcolour},   
    commentstyle=\color{codegreen},
    keywordstyle=\color{magenta},
    numberstyle=\tiny\color{codegray}, 
    stringstyle=\color{codepurple},
    basicstyle=\scriptsize\ttfamily,
    breakatwhitespace=false,         
    breaklines=true,                 
    captionpos=b,                    
    keepspaces=true,                 
    numbers=left,                    
    numbersep=5pt,                   
    showspaces=false,                
    showstringspaces=false,
    showtabs=false,                  
    tabsize=2
}

\lstset{style=mystyle}

\subsubsection{Android: Set Timer Reward}
\begin{lstlisting}[
  language=Python,                
  caption={Android Control ``Set Timer'' Generated Reward.},
  label={lst:android_reward},           
  breaklines=true,     
  numbers=left,              
  basicstyle=\scriptsize\ttfamily  
]
import json
import re
from typing import Optional, Tuple

# ------------------------------------------------------------
#               GENERIC & NORMALISATION HELPERS
# ------------------------------------------------------------

def _tab_selected(state: str, label: str) -> bool:
    pattern = (
        rf'"(content_description|text)"\s*:\s*"{label}"[^\n]*?"is_selected"\s*:\s*true'
    )
    return bool(re.search(pattern, state, re.I))

def _timer_tab_selected(state: str) -> bool:
    return _tab_selected(state, "Timer")

def _button_visible(state: str, label: str) -> bool:
    return bool(
        re.search(rf'"(content_description|text)"\s*:\s*"{label}"', state, re.I)
    )

def _is_timer_running(state: str) -> bool:
    return _button_visible(state, "Pause")

def _timer_keypad_mode(state: str) -> bool:
    return bool(re.search(r"\b\d{1,2}h\s*\d{1,2}m\s*\d{1,2}s\b", state))

# ------------------------------------------------------------
#                 TIMER / DURATION PARSING
# ------------------------------------------------------------

def _parse_requested_time(text: str) -> int:
    """Parses natural language (e.g. '5 minutes') into seconds."""
    text = text.replace("-", " ")
    hours = minutes = seconds = 0
    
    # Check explicit hours, minutes, seconds
    for patt, mult in (
        (r"(\d+)\s*hour", 3600),
        (r"(\d+)\s*minute", 60),
        (r"(\d+)\s*second", 1),
    ):
        m = re.search(patt, text, re.I)
        if m:
            val = int(m.group(1)) * mult
            if mult == 3600:
                hours = val // 3600
            elif mult == 60:
                minutes = val // 60
            else:
                seconds = val
                
    # Fallback: if "min" or raw number found
    if hours == minutes == seconds == 0:
        m = re.search(r"(\d+)\s*-?\s*min", text, re.I)
        if m:
            minutes = int(m.group(1))
        else:
            m = re.search(r"(\d+)", text)
            if m:
                minutes = int(m.group(1))
                
    total = hours * 3600 + minutes * 60 + seconds
    return total if total > 0 else 60

# ------------------------------------------------------------
#               STATE EXTRACTION
# ------------------------------------------------------------

def _extract_timer_components(state: str) -> Optional[Tuple[int, int, int]]:
    """Extracts (HH, MM, SS) from the Android UI hierarchy state."""
    
    # Case 1: Text format like "5 minutes 0 seconds"
    m = re.search(r"(\d+)\s*minutes?\s*(\d+)\s*seconds", state, re.IGNORECASE)
    if m:
        minutes = int(m.group(1))
        seconds = int(m.group(2))
        return (0, minutes, seconds)

    # Case 2: Compact format "1h 05m 20s"
    m = re.search(r"(\d+)h\s*(\d+)m\s*(\d+)s", state, re.IGNORECASE)
    if m:
        hours = int(m.group(1))
        minutes = int(m.group(2))
        seconds = int(m.group(3))
        return (hours, minutes, seconds)

    # Case 3: "MM:SS" format (excluding standard time like 12:30 PM)
    for mm_match in re.finditer(r"(\d{1,2}):(\d{2})(?!\s*[AaPp][Mm])", state):
        mm, ss = int(mm_match.group(1)), int(mm_match.group(2))
        if not (0 <= ss < 60):
            continue
        context = state[mm_match.end() : mm_match.end() + 80].lower()
        if "minute" in context or "timer" in context or "remaining" in context:
            return (0, mm, ss)

    # Case 4: Digging through UI node hierarchy if Timer tab is active
    if not _timer_tab_selected(state):
        return None

    tokens = re.findall(r'"text"\s*:\s*"([^"]+)"', state)
    tokens = [t.strip() for t in tokens]

    # Look for HH:MM:SS pattern in tokens
    for i in range(len(tokens) - 4):
        if (
            re.fullmatch(r"\d{1,2}", tokens[i])
            and tokens[i + 1] == ":"
            and re.fullmatch(r"\d{2}", tokens[i + 2])
            and tokens[i + 3] == ":"
            and re.fullmatch(r"\d{2}", tokens[i + 4])
        ):
            h = int(tokens[i])
            m_val = int(tokens[i + 2])
            s = int(tokens[i + 4])
            if 0 <= m_val < 60 and 0 <= s < 60:
                return (h, m_val, s)

    # Look for MM:SS pattern in tokens
    for i in range(len(tokens) - 2):
        if (
            re.fullmatch(r"\d{1,2}", tokens[i])
            and tokens[i + 1] == ":"
            and re.fullmatch(r"\d{2}", tokens[i + 2])
        ):
            m_val = int(tokens[i])
            s_val = int(tokens[i + 2])
            if 0 <= s_val < 60:
                return (0, m_val, s_val)

    return None

def _safe_json_dumps(obj) -> str:
    try:
        return json.dumps(obj, ensure_ascii=False)
    except Exception:
        return json.dumps({"error": "debug-serialization failed"})

# ------------------------------------------------------------
#                         REWARD CLASS
# ------------------------------------------------------------

class Reward:
    """Dense reward function specifically for setting Google Clock Timers."""

    def __init__(self, extra_info: Optional[str] = None):
        self.raw_instr: str = extra_info or ""
        self.instruction: str = self.raw_instr.lower()

        self.task_type = "set_timer"
        
        # Goal parsing
        self.goal_seconds = _parse_requested_time(self.instruction)
        rem = self.goal_seconds % 3600
        self.goal_hms = (self.goal_seconds // 3600, rem // 60, rem % 60)
        
        self.goal_achieved = False
        self._t = 0

    def reward_fn(self, state: str) -> float:
        self._t += 1
        if self.goal_achieved:
            return 100.0
        
        # Since we are only doing set_timer:
        return self._reward_timer(state)

    def debug_fn(self, state: str) -> str:
        dbg = {
            "step": self._t,
            "task_type": self.task_type,
            "goal_achieved": self.goal_achieved,
            "goal_hms": self.goal_hms,
        }
        return _safe_json_dumps(dbg)

    def _reward_timer(self, state: str) -> float:
        reward = 0.0
        
        # Check if we are on the correct tab
        if _timer_tab_selected(state):
            reward += 0.2
            
        # Extract current digits entered into the timer
        current_val = _extract_timer_components(state)
        if current_val is None:
            return min(reward, 0.99)
            
        cur_hh, cur_mm, cur_ss = current_val
        
        # Normalize to digit strings for comparison (strip leading zeros)
        current_digit_string = f"{cur_hh:02d}{cur_mm:02d}{cur_ss:02d}".lstrip("0")
        if current_digit_string == "":
            current_digit_string = "0"
            
        goal_digit_string = f"{self.goal_hms[0]:02d}{self.goal_hms[1]:02d}{self.goal_hms[2]:02d}".lstrip("0")
        if goal_digit_string == "":
            goal_digit_string = "0"
            
        # Exact match check
        if current_digit_string == goal_digit_string:
            # We treat exact digit entry as success (even if not started yet)
            # or if the timer is already running with those values.
            return 100.0
            
        # Partial match reward (dense reward for typing correct digits)
        matching_digits = 0
        for i in range(0, min(len(current_digit_string), len(goal_digit_string))):
            if goal_digit_string[i] == current_digit_string[i]:
                matching_digits += 1
            else:
                # Stop counting as soon as a mismatch occurs
                break
                
        # Scale reward based on progress
        if len(goal_digit_string) > 0:
            reward += (matching_digits / len(goal_digit_string)) * 0.7
            
        return min(reward, 0.99)
\end{lstlisting}  
\newpage

\subsubsection{MuJoCo: Hopper}
\begin{lstlisting}[
  language=Python,                
  caption={MuJoCo Hopper Generated Reward.},
  label={lst:hopper_reward},           
  breaklines=true,     
  numbers=left,              
  basicstyle=\scriptsize\ttfamily  
]
# imports
import jax
import jax.numpy as jnp

# reward function
@jax.jit
def reward_fn(state):
    reward = 0.0
    
    # Strong bonus for negative values in features 0, 1 (expert states are typically negative)
    reward += jnp.minimum(0.0, state[0]) * 2.0  # Bonus for negative feature 0
    reward += jnp.minimum(0.0, state[1]) * 2.0  # Bonus for negative feature 1
    
    # Strong penalty for positive values in features 0, 1 (learner states often positive)
    reward -= jnp.maximum(0.0, state[0]) * 2.0  # Penalty for positive feature 0
    reward -= jnp.maximum(0.0, state[1]) * 2.0  # Penalty for positive feature 1
    
    # Bonus for positive values in feature 5 (expert states are typically positive)
    reward += jnp.maximum(0.0, state[5]) * 3.0  # Bonus for positive feature 5
    
    # Penalty for negative values in feature 5 (learner states often negative)
    reward -= jnp.maximum(0.0, -state[5]) * 1.0  # Penalty for negative feature 5
    
    # Additional penalties for extreme positive values in key features
    # Features 2,3,4,6,7,10: penalize large positive values (learner indicator)
    reward -= jnp.maximum(0.0, state[2]) * 1.0
    reward -= jnp.maximum(0.0, state[3]) * 1.0
    reward -= jnp.maximum(0.0, state[4]) * 1.0
    reward -= jnp.maximum(0.0, state[6]) * 1.0
    reward -= jnp.maximum(0.0, state[7]) * 1.0
    reward -= jnp.maximum(0.0, state[10]) * 1.0
    
    # Additional penalty for large negative values in features 0,1 (to avoid over-rewarding)
    reward -= jnp.maximum(0.0, -state[0]) * 0.5
    reward -= jnp.maximum(0.0, -state[1]) * 0.5
    
    return reward
\end{lstlisting}  
\newpage

\subsection{Prompts}
\label{app:prompts}

\begin{prompt}[h!]
\centering
\begin{mymessagebox}[frametitle=Goal Identification Prompt For Data Cleaning]
\small\fontfamily{pcr}\selectfont
Given this reward code: \{reward\_code\} \\

\textbf{Trajectory:} \\
\{trajectory\} \\

Please analyze the state sequence and the agent's instruction. Identify the index of the goal state. The state indices are 1-based. \\

\textbf{OUTPUT FORMAT:} \\
Answer in a json format as follows: \\
'reasoning': Explain your reasoning for choosing the goal state(s). \\
'goal\_state\_indexes': A list of integers representing the 1-based index of the goal state(s), or -1 if no goal state is present.
\end{mymessagebox}
\caption{The prompt for identifying the goal state(s) within a trajectory.}
\label{prm:goal_identification}
\end{prompt}

\begin{prompt}[h!]
\centering
\begin{mymessagebox}[frametitle=LLM Initial Reward Generation]
\small\fontfamily{pcr}\selectfont
You are an ML engineer writing reward functions for RL training. Given a trajectory with marked goal states, create a Python reward function that can reproduce this behavior. \\

\textbf{Requirements:}
\begin{itemize}
    \item Write self-contained Python 3.9 code
    \item Make the function generic enough to handle variations (different positions, orientations, etc.)
    \item Design for modularity - you might extend this reward later to handle multiple goal types
    \item Aim to give high rewards for expert states and low rewards for all other states
\end{itemize}

\textbf{Environment Details:} \\
\{env\_code\}, \{import\_instructions\}, \{state\_description\} \\

\textbf{Trajectories} \\
\{expert\_trajectories\} \\

\textbf{Key Instructions:}
\begin{itemize}
    \item Analyze the trajectory to understand what constitutes success
    \item Identify intermediate progress that should be rewarded
    \item Create utility functions for reusable reward components
\end{itemize}

The code will be written to a file and then imported. \\
\textbf{OUTPUT FORMAT:} \\
Answer in a json format as follows: \\
'reasoning': Given the reason for your answer \\
'reward\_class\_code': Code for the Reward function class in the format: \\
\# imports \\
<imports\_here> \\
\# utils functions \\
<utils functions here> \\
\# reward function \\
class Reward: \\
\hspace*{4mm} def \_\_init\_\_(self, extra\_info=None): \\
\hspace*{8mm} <code\_here> \\
\\
\hspace*{4mm} def reward\_fn(self, state): \\
\hspace*{8mm} <code\_here> \\
\\
\hspace*{4mm} def debug\_fn(self, state): \\
\hspace*{8mm} <code\_here> \\
The Reward class will be initialized with the extra\_info argument. \\
Describe in the comments of the class the behaviour you are trying to reproduce. \\
reward\_fn and debug\_fn receive only state as argument.
The debug\_fn should return a string that will be printed and shown to you after calling reward\_fn on each state.
You can print internal class properties to help you debug the function.
Extract any needed information from the state or store it in the class.
The Reward class will be re-initialised at the beginning of each episode.
\end{mymessagebox}
\caption{Prompt to generate the initial set of rewards}
\label{prm:initial_prompt}
\end{prompt}

\begin{prompt}[h!]
\centering
\begin{mymessagebox}[frametitle=Evolution Mutation Prompt]
\small\fontfamily{pcr}\selectfont
You are an ML engineer writing reward functions for RL training. Given a trajectory with marked goal states, create a Python reward function that can reproduce this behavior. \\

\textbf{Requirements:}
\begin{itemize}
    \item Write self-contained Python 3.9 code
    \item Aim to give high rewards for expert states and low rewards for all other states
    \item Make the function generic enough to handle variations (different positions, orientations, etc.)
    \item Design for modularity - you might extend this reward later to handle multiple goal types
\end{itemize}

\textbf{Original Reward Code:} \\
\{\{code\}\} \\

\{\{import\_message\}\} \\
\{\{state\_description\}\} \\

--- \\
\textbf{CRITICAL: Incorrect Trajectories} \\
The reward function is not performing well on the following trajectories. It either assigned a high reward to a negative states or assigned low reward to an expert state. The predicted rewards for each step are shown. \\
Change the reward function to fix these errors. \\

\textbf{Key Instructions:}
\begin{itemize}
    \item Create utility functions for reusable reward components
    \item Implement goal switching logic using extra\_info to determine which reward function to use
    \item Reuse existing utilities where possible
    \item Make sure the logic you write generalises to variations in extra\_info
\end{itemize}

\{incorrect\_trajectories\} \\

\{expert\_traj\_str\} \\
--- \\

Now, provide the mutated version of the reward function that addresses these errors. \\

\textbf{OUTPUT FORMAT:} \\
Answer in a json format as follows: \\
'reasoning': Briefly explain the corrective change you made. \\
'reward\_class\_code': Code for the Reward function class in the format: \\
\# Reward format and extra info as above \\
\end{mymessagebox}
\caption{The prompt used for evolutionary mutation, providing feedback on incorrect trajectories.}
\label{prm:evolution_mutation}
\end{prompt}

\newpage




\end{document}